\documentclass[sn-mathphys-num]{sn-jnl}

\usepackage{graphicx}%
\usepackage{multirow}%
\usepackage{amsmath,amssymb,amsfonts}%
\usepackage{amsthm}%
\usepackage{mathrsfs}%
\usepackage[title]{appendix}%
\usepackage{xcolor}%
\usepackage{textcomp}%
\usepackage{manyfoot}%
\usepackage{booktabs}%
\usepackage{algorithm}%
\usepackage{algorithmicx}%
\usepackage{algpseudocode}%
\usepackage{listings}%
\usepackage{array}
\usepackage{stfloats}
\usepackage{verbatim}
\usepackage{cite}
\usepackage{stmaryrd}
\usepackage{utfsym}
\usepackage{threeparttable}
\usepackage{booktabs}
\usepackage{lmodern}
\SetSymbolFont{stmry}{bold}{U}{stmry}{m}{n}
\usepackage{anyfontsize}

\begin{document}

\title[Article Title]{Symmetrical Joint Learning Support-query Prototypes for Few-shot Segmentation}


\author[1,2]{\fnm{Qun} \sur{Li}}\email{liqun@njupt.edu.cn}

\author[1]{\fnm{Baoquan} \sur{Sun}}\email{BaoquanSun@hotmail.com}

\author*[1]{\fnm{Fu} \sur{Xiao}}\email{xiaof@njupt.edu.cn}
\author[3]{\fnm{Yonggang} \sur{Qi}}\email{qiyg@bupt.edu.cn}

\author[4]{\fnm{Bir} \sur{Bhanu}}\email{bhanu@ee.ucr.edu}

\affil[1]{\orgdiv{School of Computer Science}, \orgname{Nanjing University of Posts and Telecommunications}, \orgaddress{\street{9 Wenyuan Road}, \city{Nanjing}, \postcode{210023}, \state{Jiangsu}, \country{China}}}

\affil[2]{\orgdiv{Information Systems Technology and Design Pillar}, \orgname{Singapore University of
Technology and Design}, \orgaddress{\street{8 Somapah Road},  \postcode{487372}, \country{Singapore}}}

\affil[3]{\orgdiv{School of Artificial Intelligence}, \orgname{Beijing University of Posts and Telecommunications}, \orgaddress{\street{10 Xitucheng Road}, \city{Beijing}, \postcode{100876}, \country{China}}}

\affil[4]{\orgdiv{Department of Electrical and Computer Engineering}, \orgname{University of California at Riverside}, \orgaddress{\street{900 University Avenue}, \city{Riverside}, \postcode{92521}, \state{CA}, \country{USA}}}


\abstract{We propose Sym-Net, a novel framework for Few-Shot Segmentation (FSS) that addresses the critical issue of intra-class variation by jointly learning both query and support prototypes in a symmetrical manner. Unlike previous methods that generate query prototypes solely by matching query features to support prototypes, which is a form of bias learning towards the few-shot support samples, Sym-Net leverages a balanced symmetrical learning approach for both query and support prototypes, ensuring that the learning process does not favor one set (support or query) over the other. One of main modules of Sym-Net is the visual-text alignment based prototype aggregation module, which is not just query-guided prototype refinement, it is a jointly learning from both support and query samples, which makes the model beneficial for handling intra-class discrepancies and allows it to generalize better to new, unseen classes. Specifically, a parameter-free prior mask generation module is designed to accurately localize both local and global regions of the query object by using sliding windows of different sizes and a self-activation kernel to suppress incorrect background matches. Additionally, to address the information loss caused by spatial pooling during prototype learning, a top-down hyper-correlation module is integrated to capture multi-scale spatial relationships between support and query images. This approach is further jointly optimized by implementing a co-optimized hard triplet mining strategy. Experimental results show that the proposed Sym-Net outperforms state-of-the-art models, which demonstrates that jointly learning support-query prototypes in a symmetrical manner for FSS offers a promising direction to enhance segmentation performance with limited annotated data. The code is available at: \url{https://github.com/Sunbaoquan/Sym-Net}.}

\keywords{Symmetrical joint learning, Self-activation kernel, Visual-text alignment, Prototype-based method, Few-shot segmentation }



\maketitle

\section{Introduction}\label{sec1}
Few-Shot Segmentation (FSS) aims to achieve object segmentation in images with only a few labeled samples, such as one or five samples. Unlike traditional semantic segmentation tasks \citep{33} \citep{13}, FSS operates in a non-overlapping training and testing data paradigm. Therefore, the segmentation model should be able to learn and generalize from a limited number of samples, in a way similar to how humans learn. Specifically, the model should acquire general knowledge about to infer object categories from a few samples and be able to generalize to unseen categories. 

 Currently, most popular FSS methods are based on prototype learning~\citep{44}~\citep{06}~\citep{05}. These methods aim to extract rich support information from the limited support samples by compressing the support objects into one or multiple prototypes~\citep{06}. These prototypes are then used to match the query features, thus, segmenting the query samples. However, due to the scarcity of support images, the obtained prototypes tend to be biased towards the few-shot support samples. Therefore, the current methods overlook a crucial problem that is commonly present in few-shot tasks, which is intra-class variation. Intra-class variation arises due to factors like the viewpoint and pose changes, leading to spatial variations in the appearance of the same object class. This causes discrepancy between the support and query instances, and as a result the matching process struggles to produce optimal segmentation results. 

\begin{figure}[!t]
\centering
\includegraphics[width=1\linewidth]{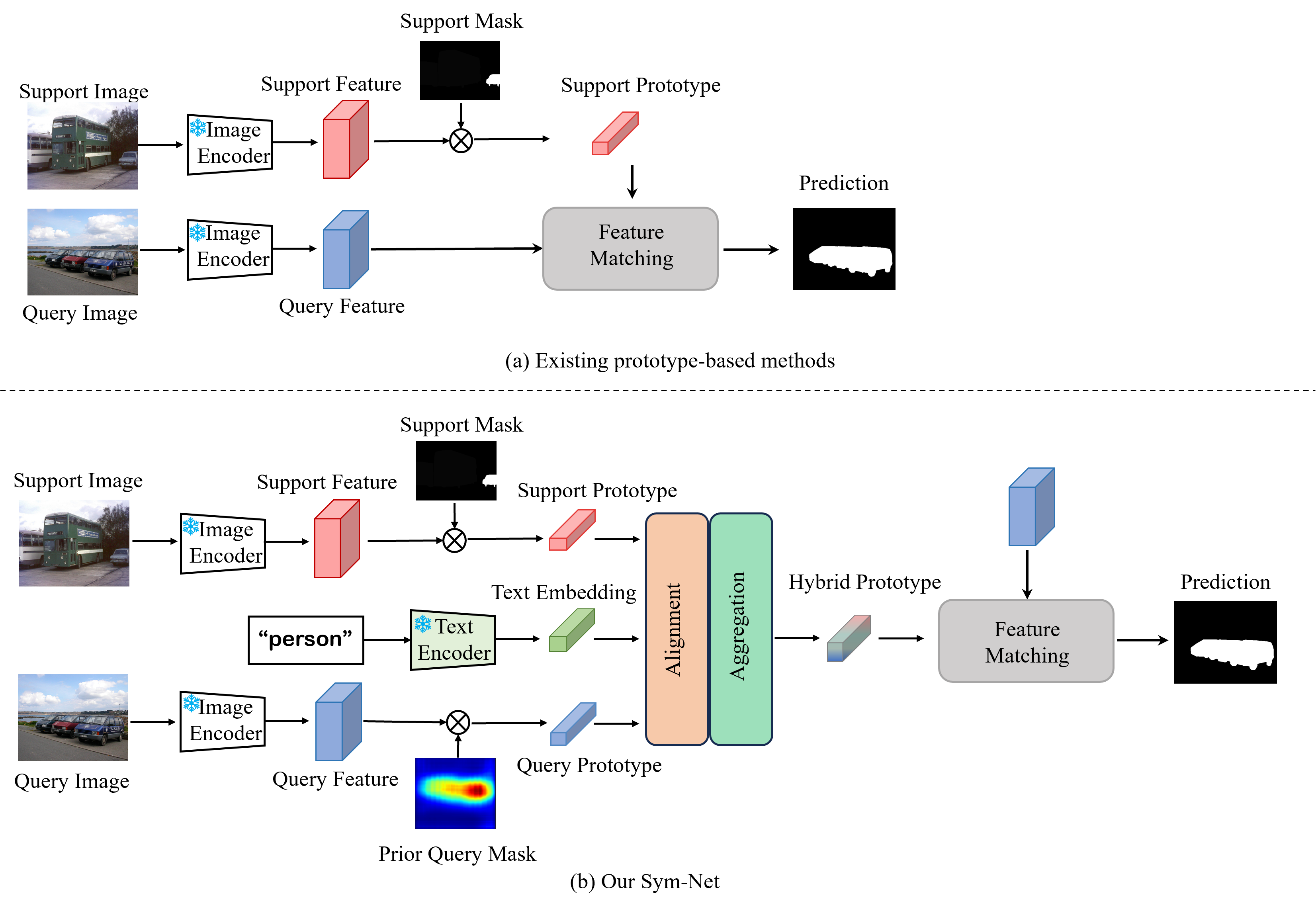}
\caption{Existing prototype-based methods vs. our proposed Sym-Net. Most of the current work (a) generates the query prototypes by performing similarity-matching the query features to the support prototypes. In contrast, our Sym-Net (b) generates the query prototypes consistent with the support prototype generation scheme in a symmetrical manner associated with the
prior query mask. The proposed method presents a query-support hybrid prototype learning
framework to match the query-support hybrid prototypes with query features.}
\label{fig_0}
\end{figure}
Considering that features from the query images could provide high-confidence complementary cues to supports, query prototypes have been explored to mitigate the intra-class variation between the query and support images \citep{05} \citep{liu2020crnet} \citep{10256677} \citep{CRCNet}. However, as shown in Fig. \ref{fig_0}(a), existing methods generate the query prototypes by simply similarity-matching the query features to the support prototypes \citep{wang2019panet}, and then either utilizing the query prototypes to match support features for auxiliary training \citep{liu2020crnet} or applying the query prototypes to refine the
obtained initial query mask \citep{05} using self-matching, which still suffer from the issue of intra-class variation due to the bias learning process.

In this paper, we seek to bridge the intra-class discrepancy by merging the object features from both the support and query images to produce more representative prototypes, as shown in Fig. \ref{fig_0}(b). We present a symmetrical query-support prototype learning framework and match the query-support hybrid prototypes with query features, which generates the query prototypes in conjunction with the prior query mask that is consistent with the support prototype generation scheme in a symmetrical manner. Specifically, a Self-activation based Prior Mask generation module (SPM) is introduced for producing a prior query mask using a self-activation kernel. We then leverage the prior query mask with high-confidence query features to construct the query prototype accordingly. In addition, a visual-text Alignment Prototype Aggregation module (APA) is proposed to to strengthen the unbiased learning of the prototypes associated by hard-positive mining. 


 Furthermore, recent research~\citep{02} has highlighted the advantages of incorporating multi-scale spatial dependencies in FSS, while the process of prototype compression in prototype-based methods eliminates the original spatial structure. Therefore, we introduce a Top-Down hyper-Correlation module (TDC) to excavate fine-grained spatial correlation at diverse scales between query and support features in a top-down manner to capture the structural cues lost due to feature compression during prototype learning.

Our major contributions are summarized as follows: 
\begin{itemize}
\item{
We propose Sym-Net, a novel framework for FSS, which tackles intra-class variation through symmetrical query-support prototype learning. Unlike prior approaches, Sym-Net achieves a balanced jointly learning of query and support prototypes, leveraging high-confidence query features within a hybrid scheme to mitigate bias in limited few-shot supports and effectively handle intra-class variations. Furthermore, Sym-Net enhances prototypes through visual-text alignment, complemented by a co-optimized hard triplet mining strategy.
}
\item{In Sym-Net, we introduce a parameter-free prior mask generation module SPM to accurately localize both local and global regions of the query object. This module utilizes sliding windows of varying sizes and a self-activation kernel to mitigate incorrect background matches effectively. Furthermore, we leverage a TDC module to inject structural cues into segmentation in a top-down manner, thereby enhancing the overall segmentation performance.}
\item{Extensive experiments demonstrate the effectiveness of Sym-Net on both PASCAL-$5^i$ and COCO-$20^i$ datasets for FSS tasks. Specifically, our approach achieves state-of-the-art results under both 1-shot and 5-shot settings, surpassing both prototype-based baseline methods and correlation-based methods. Moreover, qualitative comparisons with the second best method, highlight Sym-Net's robustness in handling scale and spatial differences between support and query images, particularly in occlusion scenarios and challenging cases.
}
\end{itemize}
\section{Related Work}
\subsection{Semantic Segmentation}
Currently, most of the existing methods \citep{13} \citep{23} \citep{32} leverage the shift-invariance of Convolutional Neural Networks (CNNs) to extract local features from images. To obtain more contextual information, various multi-scale modules and attention modules have been proposed, such as pyramid pooling~\citep{13}, atrous spatial pyramid pooling~\citep{23}, and criss-cross attention~\citep{24}. There are other methods that utilize the transformer structure~\citep{25}\citep{26} to construct global dependencies of image features. Specifically, SAM \citep{kirillov2023segany} develops a foundational model for image segmentation by employing prompt engineering. Although current segmentation methods have achieved good results, the performance of the model will severely degrade due to insufficient knowledge learning, in situations where only a few labeled samples are available for generating pixel-
level binary masks for novel class images.
\subsection{Few-shot Segmentation} 
It has been introduced to address the limitations of traditional semantic segmentation tasks, such as data dependency and weak generalization. Existing methods~\citep{31}\citep{39}\citep{40} for FSS are based on ideas from few-shot learning and metric learning to establish a metric to learn the correspondence between support and query images. 
Prototype learning~\citep{34} first applied the prototype paradigm to FSS, by compressing support images into one-dimensional prototype and utilizing it to guide the segmentation of query objects. Prototypes have played an important role  in recent work, such as in PFENet~\citep{06}, NTRENet~\citep{29} and MIANet~\citep{01}. Recognizing the limited representation capacity of a single prototype, some methods~\citep{22} have leveraged the clustering technique to obtain multiple support prototypes, aiming to improve the accuracy of feature matching with the query. To extract prototypes that are less affected by background noise, SG-One~\citep{41} has exploited a Masked Average Pooling (MAP) strategy. 
Although prototype-based methods have been found to be effective, the process of prototype compression eliminated the original spatial structure, resulting in a lack of essential spatial information.

Inherent intra-class variations have always been a challenge in FSS tasks. When there are significant appearance changes between the support and query objects, relying solely on weakly correlated support information to guide query object segmentation has led to poor results. There have been visual correspondence-based methods~\citep{02} \citep{10} \citep{nie2024crossdomain} that establish visual correspondence between the support and query sets, and prompt-based methods~\citep{zhu2024llafs} that leverage the prior
knowledge gained by large language model, achieving notable results albeit with complex designs and high computational cost. We have focused on prototype learning, but unlike previous methods, our prototype representation has integrated object-relevant features from both query and support images to learn hybrid prototypes, enabling it to cover a wider range of variations in target objects beyond the support set. Even when there are significant differences between support and query instances, our model has been found to robustly segment the query objects.
\subsection{Semantic Information for Segmentation}
Recent methods have attempted to use multimodal data for segmentation tasks, including embedding semantic information by using text labels. On the one hand, text labels have been easier to obtain as compared to the manually annotated masks, and which has reduced the annotation workload significantly. On the other hand, text data, when encoded with pre-trained Natural Language Processing (NLP) models, carried rich descriptive semantic information. Specifically, CLIPSeg~\citep{20} utilized the pre-trained CLIP~\citep{27} as the backbone to generate masks of target objects based on text prompts. 
IMR-HSNet~\citep{21} obtaind class activation mapping by feeding the class-level label of query images into the text encoder of CLIP, achieving weakly supervised FSS. MIANet~\citep{01} incorporated class label as the semantic information into the support set by using a general information module. Considering the significance of semantic information in segmentation, we attempted to leverage semantic guidance to provide prior knowledge for prototype learning.
\begin{sidewaysfigure*}[thbp]
\centering
\includegraphics[width=1\linewidth]{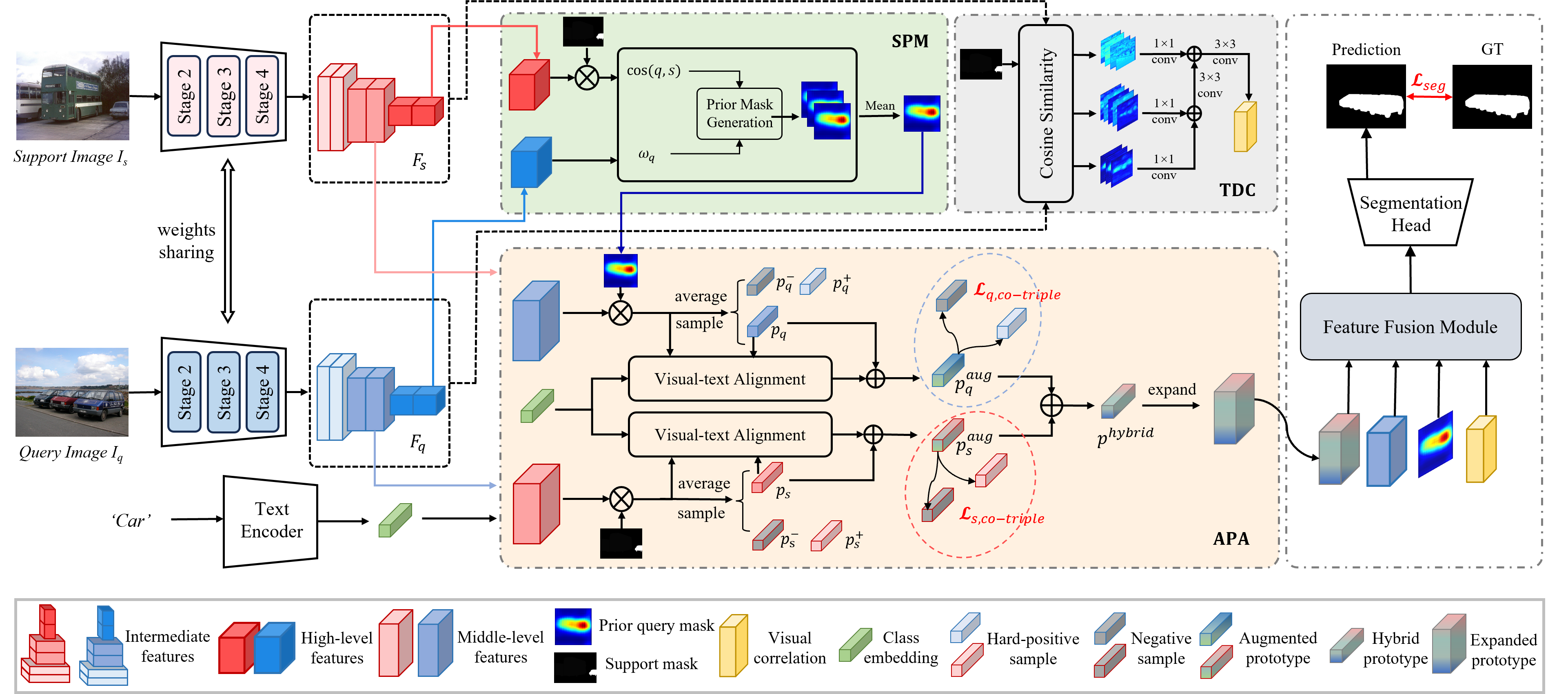}
\caption{Overview architecture of Sym-Net in 1-shot setting. Sym-Net consists of three main modules, including a visual-text Alignment based Prototype Aggregation module (APA), a Top-Down hyper-Correlation module (TDC) and a Self-activation based Prior Mask generation module (SPM), which are shadowed in color of orange, gray and cyan, respectively.
First, we extract intermediate features ${F}_s$ and ${F}_q$ for both the support image $\mathbf{I}_s$ and the query image $\mathbf{I}_q$ by employing a weight-share feature extractor, and extract text embeddings for the class label by using text encoder. Then, we generate informative prototypes, $\mathbf{p}_q^{aug}$ and $\mathbf{p}_s^{aug}$, in a symmetrical manner by using the visual-text alignment combined with hard-positive mining trick in the APA module. Specifically, the SPM module is introduced to generate the prior mask for query image. In addition, we design the TDC module to compute the visual hyper-correlation features $\mathbf{F}^{hyper}$. After that, all features that are obtained above, including the prototypes, the visual features, the prior mask and the visual hyper-correlation features, are fed into the feature fusion module to generate the predicted query mask $\hat{\mathbf{M}}_q$ associated by a segmentation head. For the $K$-shot setting, where $k$ support samples are available, we calculate $K$ prior masks, $K$ support prototypes, and $K$ sets of correlation features, and average them.}
\label{fig_1}
\end{sidewaysfigure*} 
\section{Technical Approach}
\subsection{Problem Formulation}

We adopt an episode-based meta-learning approach to tackle the issue of data scarcity in FSS tasks \citep{01, 05}. We define two datasets, $D_{train}$ and $D_{test}$, representing the training set and test set, respectively. These datasets consist of multiple episodes randomly sampled from the training classes $C_{train}$ and the test classes $C_{test}$. Importantly, the categories in $C_{train}$ and $C_{test}$ do not overlap $C_{train} \cap C_{test} = \emptyset$. The model undergoes training on $D_{train}$ and is subsequently evaluated on $D_{test}$. Following \citep{01}, all class labels are mapped to text embedding vectors, denoted as $(\mathbf T_{train},\mathbf T_{test}) = \mathcal{F}(C_{train}, C_{test})$. In the $K$-shot scenario, each episode comprises a support set $S$ and a query set $Q$, accompanied by the text embedding $\mathbf t$ of the sampled classes, denoted as $E= (S, Q, \mathbf t)$.
The support set $S = \{<\mathbf{I}_s^i, \mathbf{M}_s^i>\}_ {i=1}^K$ consists of $K$ pairs of support samples, where $\mathbf{I}_s^i$ represents the $i$-th support image and $\mathbf{M}_s^i$ represents its corresponding mask. The query set $Q = <\mathbf{I}_q, \mathbf{M}_q>$ contains only one query image $\mathbf {I}_q$ and its ground-truth mask $\mathbf{M}_q$. Given the query image $\mathbf{I}_q$, the goal is to utilize $(S, \mathbf t)$ to obtain its predicted mask $\hat{\mathbf M}_q$.

\subsection{Model Overview}
As shown in Fig.~\ref{fig_1}, we present the overview architecture of Sym-Net, which learns the query and support prototypes in a symmetrical manner. Sym-Net consists of three main modules, including the APA, TDC and SPM, which are shadowed in color of orange, gray and cyan, respectively. First, we extract the support feature set $F_s=\mathbf{F}_s^h \cup \mathbf{F}_s^m \cup \mathbf{F}_s^l$ and the query feature set $F_q=\mathbf{F}_q^h \cup \mathbf{F}_q^m \cup \mathbf{F}_q^l$ by employing a weight-sharing feature extractor for both the support image $\mathbf I_s$ and the query image $\mathbf I_q$, where $\mathbf{F}_s^h,\ \mathbf{F}_s^m,\ \mathbf{F}_s^l$ represents high-level support features, middle-level support features and low-level support features, and $\mathbf{F}_q^h,\ \mathbf{F}_q^m,\ \mathbf{F}_q^l$ represents high-level query features, middle-level query features and low-level query features, respectively. We also extract text embeddings for the class label using a text encoder. Next, the middle-level support and query features $\mathbf{F}_s^m$ and $\mathbf{F}_q^m$, along with the text embedding $\mathbf t$ are input to the visual-text alignment module assisted by the hard-positive mining trick to generate informative prototypes $\mathbf{p}^{aug}$ in the APA module. Specifically, the prior mask for the query image is obtained from the SPM module using a self-activation kernel. Additionally, we design the TDC module to obtain the visual hyper-correlation features $\mathbf{F}^{hyper}$. Afterwards, all obtained features, including the prototype, the prior mask, 
and hyper-correlation features, are fed into the feature fusion module, where they can interact with each other. Finally, a segmentation head is employed to predict the query mask $\hat{\mathbf{M}}_q$. 

\subsection{Self-activation based Prior Mask Generation Module (SPM)} 

Previous methods \citep{01,06} obtain the prior mask for the query image by using extracted high-level features from the support and query sets. This prior mask provides a rough indication of the probability that each pixel in the query image belongs to the target class. However, the pixel-wise calculation of the prior mask ignores local correlations, leading to erroneous matches in certain regions. For example, the generated prior mask may have high activation values in regions that do not belong to the query object. Therefore, the obtained prior mask is not entirely reliable for locating the query object, which inevitably has a negative impact on subsequent steps. We design the SPM module to generate the prior mask for the query image without introducing learnable parameters while accurately localizing both local and global regions of the query object. We employ sliding windows of different sizes to extract local representations from the high-level support and query features, allowing us to establish correspondences between the support and query images in different ways. When computing region-based similarity, we incorporate self-activation kernel to suppress incorrect matches between background pixels in the query and the support objects.

The SPM module takes high-level support and query features, i.e., $\mathbf{F}_s^h\in \mathbb{R}^{H\times W\times C}$ and $\mathbf{F}_q^h\in \mathbb{R}^{H\times W \times C}$, and the corresponding support mask $\mathbf{M}_s\in \mathbb{R}^{\hat{H}\times \hat{W}}$ as inputs. Here, $C$ represents the number of the channels, $H$ and $W$ denote the resolution of the features, and $\hat{H}$ and $\hat{W}$ are the height and width of the support mask, respectively. Initially, average pooling with distinct window sizes is performed on both the support and query samples to acquire their region features, i.e., $R_{s}$ and $R_{q}$. There are no trainable parameters introduced in this process, hence it is beneficial for preventing over-fitting to the seen classes. 
Particularly, the support mask $\mathbf{M}_s$ is utilized to exclude the background of support samples, and padding is applied during the average pooling  to ensure consistency in the output size. In formal, the region feature $r_{s}\in R_{s}$ and $r_{q}\in R_{q}$ are defined by: 
\begin{eqnarray}
\label{equ:1}
    r_{s}&=& \mathcal{F}_{{AvgPool}\ (d_{H} \times d_{W})}\left( \mathbf{F}_s^h\otimes\varphi\left(\mathbf{M}_{s} \right) \right),\nonumber \\
r_{q}&=& \mathcal{F}_{{AvgPool}\ (d_{H} \times d_{W})}\left( \mathbf{F}_{q}^{h} \right),
\end{eqnarray}
where $\mathcal{F}_{{AvgPool}\ (d_{H} \times d_{W})}(\cdot)$ represents average pooling with the window size of $d_{H} \times d_{W}$, and $\varphi(\cdot)$ denotes the bilinear interpolation, utilized to map $\mathbf{M}_s$ to the same size as $\mathbf{F}_s^h$. The symbol $\otimes$ denotes the Hadamard product. 

\begin{figure}[!t]
\centering
\includegraphics[width=1\linewidth]{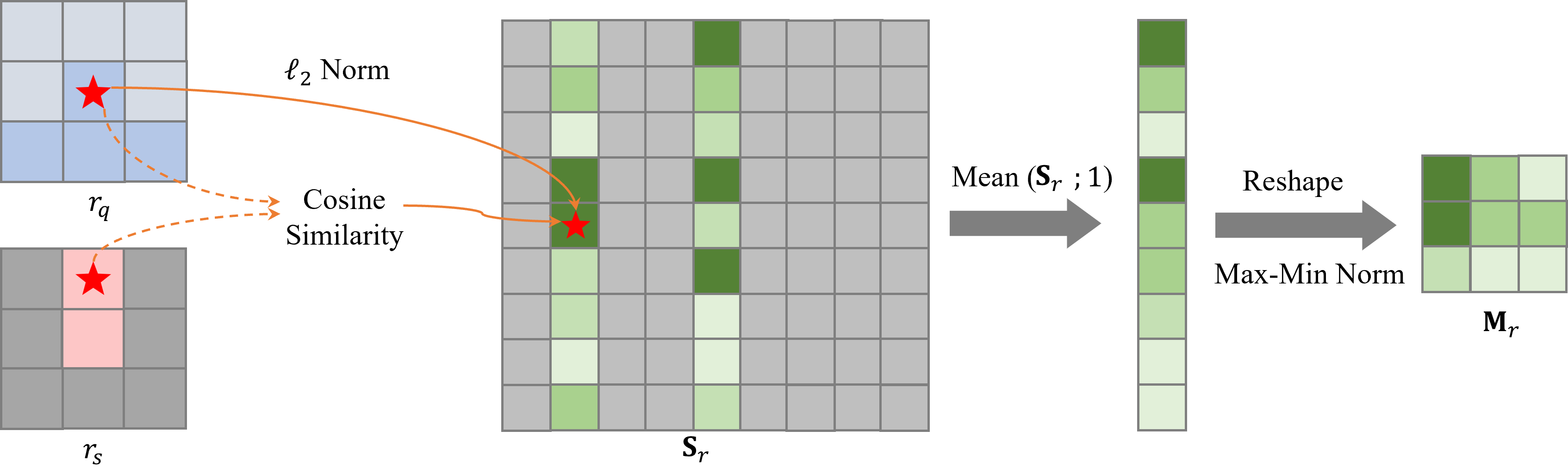}
\caption{Schematic overview of Self-activation based Prior Mask generation (SPM). Given a pair of region feature $r_q$ and $r_s$, we obtain a region-based matching score $\mathbf S_r$ after performing region-wise similarity calculations based on a self-activation kernel. By doing so,
we can obtain a region-based affinity matrix $\mathbf S_r$. Subsequently, we calculate the average of $\mathbf S_r$ along the second dimension and employ min-max normalization to reshape the feature map to obtain the region-based similarity map $\mathbf{M}_{r}$. $Mean(S_{r};1)$ denotes the average along the second dimension of $\mathbf S_r$. 
}
\label{fig_2}
\end{figure}
Following this, we conduct region-based matching between the query and support region features. As shown in Fig.~\ref{fig_2}, we introduce a self-activation kernel $\omega_q$, which can be interpreted as region-based norms, to re-weight the matching scores between the query and support features. This is to mitigate incorrect matches that may arise due to complex query backgrounds. Therefore, for each pair of region feature $(r_q, r_s)$, their matching score $s_r\in \mathbb{R}$ is formed as follows:
\begin{eqnarray}
\label{equ:2}
    s_r &=&\cos\left( r_{q},r_{s} \right) \cdot  \omega_{q}, \nonumber \\
    &=&\cos\left( r_{q},r_{s} \right) \cdot < r_{q}, r_{q} >,
\end{eqnarray}
where $\omega_{q}= < r_{q}, r_{q} >$ is the self-activation kernel, which can be obtained as the L2-norm of $r_{q}$, i.e., $\| r_{q}\|_2$. 
By doing so,
we can obtain a region-based affinity matrix $\mathbf S_r \in \mathbb{R}^{HW\times HW}$, with the first dimension corresponding to the query and the second dimension associated with the support. Subsequently, we calculate the average of $\mathbf S_r$ along the second dimension and employ min-max normalization to reshape the feature map to the same spatial size $H\times W$ as the query feature. This process results in the creation of the region-based similarity map $\mathbf{M}_{r}\in \mathbb{R}^{H \times W}$, which is formally defined as:
\begin{eqnarray}
\label{equ:3}
    \mathbf{M}_{r} = \mathcal {F}_{reshape}\left( Norm\left( Mean\left( \mathbf{S}_{r};1 \right) \right) \right),
\end{eqnarray}
where $Mean(S_{r};1)$ represents
the average of $\mathbf S_r$ along the second dimension. 
Finally, the prior mask $\mathbf{M}^{pri}$ for the query image is obtained by averaging all the region-based similarity maps.


\subsection{Visual-text Alignment based Prototype Aggregation Module (APA)} 
Existing prototype-based methods~\citep{01, 06} extract single or multiple prototype representations of the target class by using support features and support masks. These methods have shown excellent performance in handling highly correlated support-query sample pairs. However, FSS tasks are challenged by intra-class variations. When substantial deviations exist between the query and support objects, the support prototypes may struggle to match those deviated regions in the query, leading to sub-optimal results. In addition to extracting support prototypes, the proposed APA module also captures query prototypes from the query set and semantic information from the class label. It then adaptively combines these pieces of information. The resulting prototype representations exhibit class-specific characteristics, facilitating accurate pixel classification similar to the target class during the matching process with query pixels. As a result, the APA module enhances the model's capability to address intra-class variations. It is noteworthy that the dual-branch scheme employed for generating the support and query prototypes in the APA module follows a symmetrical pattern.  
\subsection{Details of APA}
As depicted in Fig.~\ref{fig_1}, the APA module comprises three input branches, including the query, the semantic embedding, and the support branch. Essentially, the semantic embedding of each class possesses a general feature anchor for the category, offering prior knowledge. First, we employ a pre-trained NLP model to extract the semantic information of the class. Then, the support and query branches are individually fused with the semantic information by visual-text alignment, resulting in enhanced support and query prototypes, respectively. Finally, these two enhanced prototypes are integrated in a weighted manner. We expand upon the triplet loss introduced by MIANet~\citep{01} by implementing a co-optimized hard triplet mining strategy. This modification aims to jointly optimize the module, enhancing its overall effectiveness.

Specifically, we initiate the process by inputting class labels into a pre-trained NLP model to acquire the semantic embedding $\mathbf t \in \mathbb{R}^{1\times d}$, where $d$ is the dimension of the semantic embedding. Next, for the support branch, we apply standard MAP~\citep{41} to the support mask $\mathbf{M}_s$ and middle-level support feature $\mathbf{F}_s^m$ to derive the support prototype $\mathbf p_{s}\in \mathbb{R}^{1 \times 1 \times c}$ as:
\begin{eqnarray}
\label{equ:4}
    \mathbf p_{s} = \frac{\sum_{i = 1}^{HW}{\mathbf{F}_{s}^m(i)} \cdot \llbracket\varphi( \mathbf{M}_{s})(i) = 1\rrbracket}{ \sum_{i = 1}^{HW}\llbracket\varphi\left(\mathbf{M}_{s}\right)(i) = 1\rrbracket}, 
\end{eqnarray}
where $\llbracket\cdot\rrbracket$ is the indicator function,  if $\varphi\left(\mathbf{M}_{s}\right)(i)=1$ is true, the value of $\llbracket\varphi\left(\mathbf{M}_{s}\right)(i)=1\rrbracket$ equals to 1, otherwise it is set to 0. $c$ corresponds to the channel dimension.

For the query branch, due to the absence of the query mask, we leverage the prior mask predicted by SPM to derive the
query prototype $\mathbf p_{q}\in \mathbb{R}^{1 \times 1 \times c}$, which is defined as follows:
\begin{eqnarray}
\label{equ:5}
   \mathbf p_{q} = \frac{\sum_{i = 1}^{HW}{\mathbf{F}_{q}^m(i)} \cdot \llbracket \mathbf M^{pri}(i)>\tau_1\rrbracket}{\sum_{i = 1}^{HW}\llbracket\mathbf M^{pri}(i)>\tau_1\rrbracket},
\end{eqnarray}
where $\tau_1$ signifies the threshold for $\mathbf M^{pri}$, governing the filtering range of the foreground in the query feature. If the condition $\mathbf{M}^{pri}(i)>\tau_1$ holds true, the value of $\llbracket\mathbf{M}^{pri}(i)>\tau_1\rrbracket$ is set to 1, and 0 otherwise. Then, we concatenate $\mathbf p_q,\ \mathbf p_s$ with $\mathbf t$ to obtain their enhanced representation $\mathbf p_{q,t},\ \mathbf p_{s,t}$, respectively, which are formulated as:
\begin{equation}
\label{equ:6}
   \begin{cases}
   \mathbf p_{q,t}=\mathbf p_q\oplus \mathbf t, \ \mathbf p_{q,t} \in \mathbb{R}^{1\times 1\times (c+d)},\\   
   \mathbf p_{s,t}=\mathbf p_s\oplus \mathbf t,  \ \mathbf p_{s,t} \in \mathbb{R}^{1\times 1\times (c+d)},
   \end{cases}
\end{equation}
where $\oplus$ represents the operation of concatenation along the channel dimension. Subsequently, they are fed into the visual-text alignment module. 

Given the similar process of the visual-text alignment for both the query and support branches, we here illustrate the details of this module using the query branch.
During visual-text alignment, we align visual features with semantic embedding to acquire a prototypical representation that establishes a correspondence between visual and language.
In particular, the query $\mathbf Q$ is derived from $\mathbf p_{q,t}$, while the key $\mathbf K$ and value $\mathbf V$ are extracted from the weighted query features $\widetilde{\mathbf{F}_q}=\mathbf M^{pri}\times \mathbf{F}_q^m \in \mathbb{R}^{H\times W\times C}$, as formulated below: 
\begin{equation}
\label{equ:7}
    \mathbf Q = \mathbf p_{q,t}\mathbf W^{Q},\ 
    \mathbf K = \widetilde{\mathbf{F}_{q}} \mathbf W^{K},\ 
    \mathbf V = \widetilde{\mathbf{F}_{q}} \mathbf W^{V},
\end{equation}
where $\mathbf W^ Q\in \mathbb{R}^{(c+d)\times c}$ is the linear projection layer, and $\mathbf W^K$ and $\mathbf W^V$ refer to $1\times 1$ convolutions with an output channel of $c$. Then, we reshape $\mathbf K$ and $\mathbf V$ to the dimension of $HW\times c$ to calculate standard single-head attention. After that, we can derive the augmented query prototype $\mathbf p_q^{aug}\in \mathbb{R}^{1 \times 1 \times c}$ as:
\begin{equation}
\label{equ:8}
    \mathbf p_{q}^{aug} = FFN(Softmax\left(\frac{\mathbf Q\mathbf K^{T}}{\sqrt{d}} \right)\mathbf V) + \mathbf p_{q}, 
\end{equation}
where $d$ denotes the scaling factor. 

\begin{figure}[!t]
\centering
\includegraphics[width=0.4\linewidth]{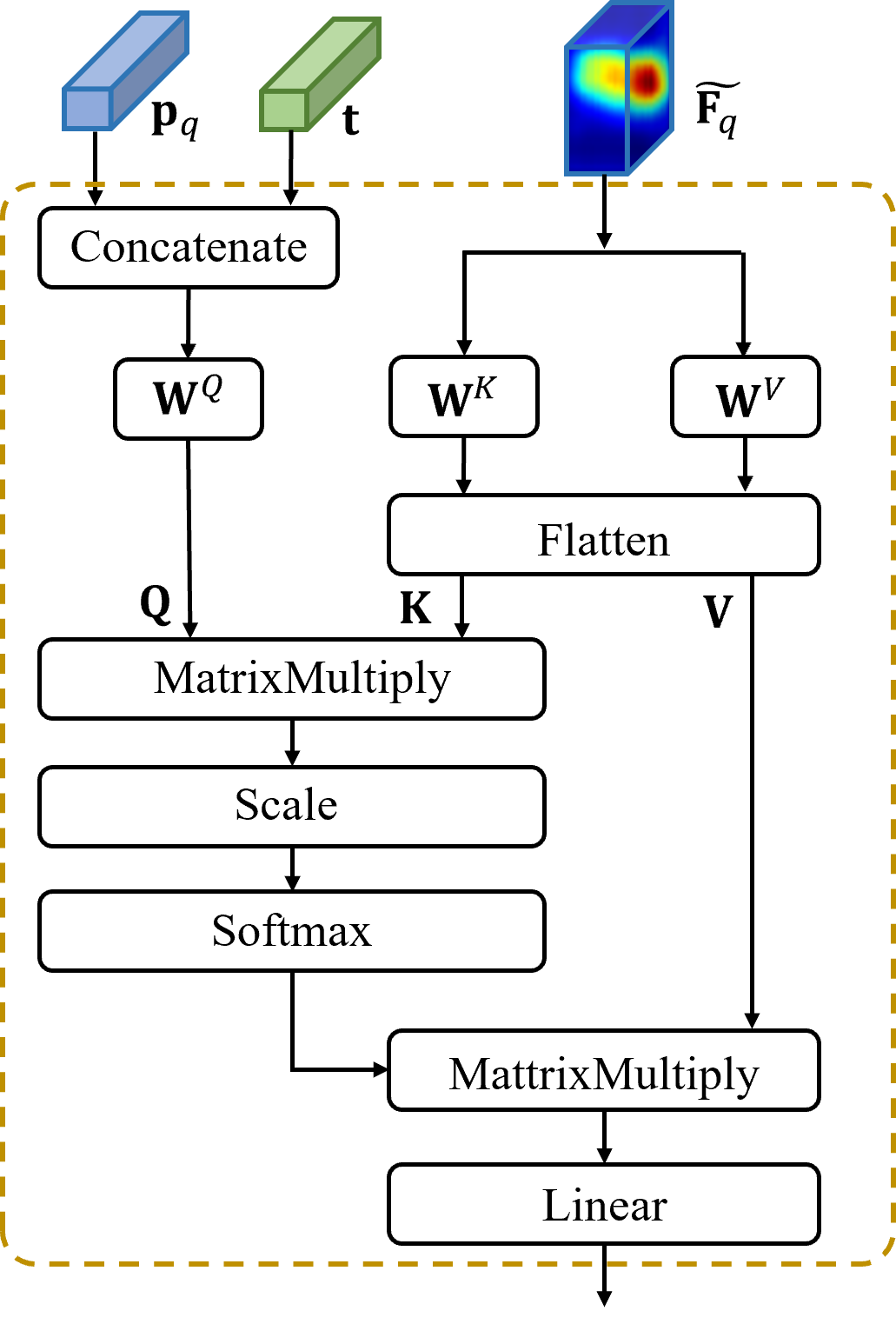}
\caption{Architecture of visual-text Aggregation based Prototype Alignment (APA) module for the query branch.}
\label{sup1}
\end{figure}

 Fig. \ref{sup1} presents the architecture of APA for the the query branch as an example. First, this module concatenates the query prototype $\mathbf p_q$ with the semantic embedding $\mathbf t$ to obtain the concatenated representation $\mathbf p_{q,t}$ as defined in Equation (\ref{equ:6}), and then generates the query $\mathbf Q$ from $\mathbf p_{q,t}$, the key $\mathbf K$ and value $\mathbf V$ from the weighted query features $\widetilde{\mathbf{F}_q}=\mathbf M^{pri}\times \mathbf{F}_q^m$ by using the $\mathbf W^Q$, $\mathbf W^K$ and $\mathbf W^V$, respectively, as defined in Equation (\ref{equ:7}). Finally, this module conducts standard single-head attention. APA aligns visual features with semantic embedding to acquire a prototypical representation that establishes a correspondence between visual and language. Different from cross-attention, it aggregates text and visual features by concatenating them directly.

In the same way, we can obtain the augmented support prototype $\mathbf p_s^{aug}\in \mathbb{R}^{1 \times 1 \times c}$ from the support branch. Furthermore, we combine $\mathbf p_s^{aug}$ and $\mathbf p_q^{aug}$ with balancing parameters $\alpha$ and $\beta$ to consolidate the augmented hybrid prototype $\mathbf{p}^{aug}\in \mathbb{R}^{1 \times 1 \times c}$ corresponding to a specific class:
\begin{equation}
\label{equ:9}
    \mathbf{p}^{hybrid} = \alpha \mathbf p_{q}^{aug} + \beta \mathbf p_{s}^{aug}.
\end{equation}



\subsection{Query-support Joint Loss for Hard Triplet Mining} During prototype learning, a query-support joint loss for hard triplet mining is developed to enhance the ability of the unified class prototype to distinguish the foreground from the background.
We initiate the process by sampling positive-negative pairs in both the query and support branches. Concretely, within the query branch, $\mathbf p_q^{aug}$ acts as the anchor, and positive and negative samples are determined by aggregating query features in the guidance of the prior mask $\mathbf M^{pri}$.
As the background is cluttered and minimally relevant to the query object, we consolidate regions with low activation values in query features into a unified background prototype, forming the negative query sample $\mathbf p_{q}^{-}\in \mathbb{R}^{1 \times 1\times c}$:
\begin{equation}
\label{equ:10}
    \mathbf p_{q}^{-} = \frac{\sum_{i = 1}^{HW}\mathbf{F}_{q}^m(i)\cdot \llbracket\mathbf M^{pri}(i) < \tau_{2}\rrbracket}{\sum_{i = 1}^{HW}\llbracket\mathbf M^{pri}(i) < \tau_{2}\rrbracket},
\end{equation}
where the threshold $\tau_2$ is used to regulate the selection range of negative samples for query features. To construct positive feature samples, we utilize the hardest ones as positive samples akin to MIANet \citep{01}. Intuitively, the boundary of the query object tends to be ambiguous, making it uncertain whether these pixels belong to the foreground or background, namely the hardest samples. To utilize these hard foreground samples, we only include pixels where the activation probability of the query feature falls within the intermediate range,
thus aggregating them as the hard-positive query sample $\mathbf p_{q}^{+}$ by the following:
\begin{equation}
\label{equ:11}
    \mathbf p_{q}^{+} = \frac{\sum_{i = 1}^{HW}\mathbf{F}_{q}^m(i)\cdot \llbracket\tau_{3}<\mathbf M^{pri}(i) < \tau_{4}\rrbracket}{\sum_{i = 1}^{HW}\llbracket\tau_{3}<\mathbf M^{pri}(i) < \tau_{4}\rrbracket},
\end{equation}
where $\tau_3,\ \tau_4$ is used to manage the selection range of hard-positive samples for query features. 

In the support branch, $\mathbf p_s^{aug}$ serves as the anchor. Guided by support masks, the negative support sample $\mathbf p_s^-$ is obtained by averaging the background of the support features, similar to the calculation of $\mathbf p_q^-$. Similarly, we select the farthest foreground pixel from $\mathbf p_s^{aug}$ as the hard-positive support sample $\mathbf p_s^+$. Therefore, the joint triplet loss can be defined as follows:
\begin{flalign}
\label{equ:12}
    \mathcal{L}_{co-triple}=max( \parallel \mathbf p_{q}^{aug} - \mathbf p_{q}^{+}\parallel + 0.5-\parallel \mathbf p_{q}^{aug} - \mathbf p_{q}^{-}\parallel, 0 )+\nonumber\\
max( \parallel \mathbf p_{s}^{aug} - \mathbf p_{s}^{+}\parallel + 0.5 - \parallel \mathbf p_{s}^{aug} - \mathbf p_{s}^{-}\parallel,0).
\end{flalign}

\subsection{Top-down Hyper-Correlation Module (TDC)} 
To address the spatial details lost due to prototype spatial pooling, we incorporate a TDC module to amalgamate hierarchical visual correlations and extract fine-grained features.
First, we compute the cosine similarity between pair-wise features from the support feature set $F_s$ and the query feature set $F_q$ to obtain a set of correlation maps $Corr$. Since $F_s$ and $F_q$ are obtained from the second to fourth stages of the backbone network, we split $Corr$ into three different levels. Then, we concatenate correlation maps belonging to the same level along the channel dimension, resulting in three correlation features $\mathbf{F}_{corr1}\in \mathbb{R}^{H\times W\times N_1}$, $\mathbf{F}_{corr2}\in \mathbb{R}^{H\times W\times N_2}$, and $\mathbf{F}_{corr3}\in \mathbb{R}^{H\times W \times N_3}$, where $N_1+N_2+N_3=N$. Next, we feed $\mathbf{F}_{corr1}, \mathbf{F}_{corr2}, \mathbf{F}_{corr3}$ into a top-down multi-scale structure to fuse them gradually in a top-down manner. Finally, the hyper-correlation features $\mathbf{F}^{hyper}\in \mathbb{R}^{H\times W\times N^{'}}$ is obtained by: 
\begin{small}
\begin{flalign}
\label{equ:13}
        &\mathbf{F}^{hyper}= \mathcal {F}_{Conv_{3\times 3}}\Big(\mathcal {F}_{Cat} \big[\mathcal {F}_{Conv_{3\times 3}} \big(\mathcal {F}_{Cat}[\mathcal {F}_{Conv_{1\times 1}}(\mathbf{F}_{corr3}),\nonumber\\
        & \mathcal {F}_{Conv_{1\times 1}} (\mathbf{F}_{corr2})]\big),\mathcal {F}_{Conv_{1\times 1}}(\mathbf{F}_{corr1})\big]\Big),
\end{flalign}
\end{small}
where $\mathcal {F}_{Cat}$ denotes feature concatenation. $\mathcal {F}_{Conv_{1\times 1}}$ and $\mathcal {F}_{Conv_{3\times 3}}$ are $1\times1$ and $3\times 3$ convolutions.
\subsection{Training Objective}
In the following, we describe how to utilize the obtained $\mathbf M^{pri}$, $\mathbf{p}^{aug}$, and $\mathbf{F}^{hyper}$
to guide the segmentation of the query image. We opt for the Feature Enrichment Module (FEM) proposed in PFENet \citep{06} as our feature fusion module, namely the feature decoder. This module takes $\mathbf{F}_q$, $\mathbf M^{pri}$, $\mathbf{p}^{aug}$, and $\mathbf{F}^{hyper}$ as inputs.
First, $\mathbf{p}^{aug}$ is expanded to match the dimension of other features. Then, all features are either interpolated or adaptively pooled and fed into the FEM module to generate the intermediate prediction at four scales $\{\hat{\mathbf M}_{inter}^{k}\}_{k=1}^4$ and the ultimate predicted query mask $\hat{\mathbf M}_q$. 

Therefore, the entire loss consists of the joint triplet loss $\mathcal{L}_{co-triple}$ as shown in Equation (\ref{equ:12}) and the segmentation loss $\mathcal L_{seg}$. Specifically, $\mathcal L_{seg}$ comprises two components: $\mathcal L_{inter-seg}$, which measures the difference between the intermediate prediction results $\{\hat{\mathbf M}_{inter}^{k}\}_{k=1}^4$ and the ground-truth of the query mask $\mathbf{\mathbf M}_q$, and $\mathcal L_{final-seg}$, which quantifies the disparity between the final prediction result $\hat{\mathbf M}_q$ and the ground-truth of the query mask $\mathbf{\mathbf M}_q$. Both losses are computed using the Binary Cross-Entropy (BCE) function:
\begin{flalign}
\label{equ:14}
    &\mathcal L_{seg} =\mathcal L_{inter-seg} + \mathcal L_{final-seg}\nonumber \\
    &=\frac{{\sum_{k=1}^4}{{\frac{1}{HW}}\sum_{i=1}^H\sum_{j=1}^{W}BCE\big(\hat{\mathbf M}_{inter}^k(i,j),\mathbf{M}_q(i,j)\big)}}{4}\nonumber \\
    &+\frac{\sum_{i=1}^H\sum_{j=1}^{W}BCE\big(\hat{\mathbf M}_{q}(i,j),\mathbf{M}_q(i,j)\big)}{HW}.
\end{flalign}

Therefore, the total loss is defined by:
\begin{equation}
\label{equ:15}
    \mathcal L = \mathcal L_{co-triple} + \mathcal L_{inter-seg} + \mathcal L_{final-seg}.
\end{equation}


\begin{figure*}[!ht]
\centering
\includegraphics[width=1\linewidth]{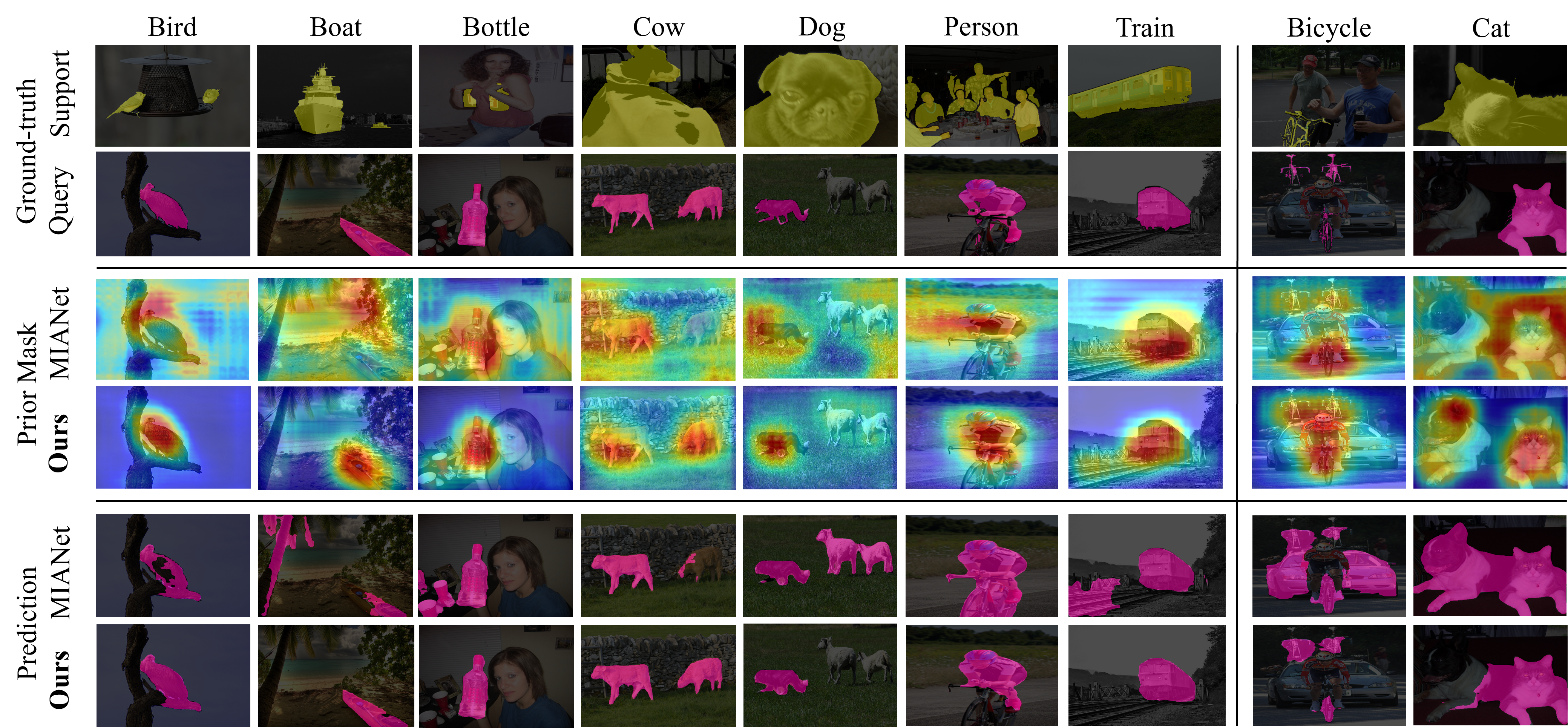}
\caption{Visualization of the prior query mask and segmentation of Sym-Net (ours) vs. MIANet. The prior mask provides a preliminary indication of the approximate position of the query object. 
Benefiting from the prior query mask generated by the self-activation based prior mask generation module (SPM), more accurate segmentation is achieved, even in hard cases. Note that we average four-scale masks from MIANet for a fair comparison. Best viewed in color.
}
\label{fig_3}
\end{figure*}
\section{Experimental Results and Analysis}
\subsection{Experimental Details} 
\noindent{\textbf{Datasets.}} We conduct experiments on PASCAL-$5^i$ and COCO-$20^i$ datasets. The PASCAL-$5^i$ dataset \citep{35} with $20$ object classes is created by additional annotations from PASCAL VOC 2012~\citep{14} and SBD~\citep{14}. The COCO-$20^i$ dataset \citep{16} with $80$ object classes is more challenging, which is constructed from the MSCOCO dataset~\citep{17}. We divide all categories into four folds by following the same split as in~\citep{06} \citep{01} for 4-fold cross-validation. As a result, each fold contains 5 categories on the PASCAL-$5^i$ dataset, and 20 categories on the COCO-$20^i$ dataset, respectively. In each experiment, three folds are used for training, and the remaining fold for testing.\\
\textbf{Metric and Evaluation.} Following previous methods \citep{06}\citep{01}, we use mean Intersection over Union (mIoU) as the metric to measure the performance of our model. We perform 5 rounds of testing with 5 different random seeds and report the average results. We randomly sample 1000 support-query pairs from the PASCAL-$5^i$ dataset and 5000 pairs from the COCO-$20^i$ dataset in each testing round, which are consistent with previous methods~\citep{01}.\\
\textbf{Implementation Details.} 
In default baseline set up, we utilize pretrained ResNet50 \citep{11} as the feature extractor and Word2Vec \citep{word2vec} as the text encoder, and their parameters are frozen during training. It should be noted that we choose feature maps of all blocks from the second to fourth stages in ResNet50 as the low-level features, middle-level features and high-level features, respectively. The dimensions of features that are extracted by ResNet50 is 512, 1024 and 2048 from low to high. For obtaining text embeddings of a class, we use the pretrained NLP model, Word2Vec \citep{word2vec}, which outputs text embeddings with 300 dimensions. Following the works in~\citep{01, 10} and \citep{09}, PSPNet~\citep{13} serves as the base learner, while our entire model acts as the meta learner. We train for 200 and 50 epochs in a batch size of 8 for all models on the PASCAL-$5^i$ dataset and COCO-$20^i$ dataset, respectively. During training, we use the Stochastic Gradient Descent (SGD) optimizer with a learning rate of 0.005. The weight decay is set to 0.0001, and the momentum is set to 0.9. 
In our experiments, we use three pooling windows in different directions, including $d_{h} \times d_{w}=\{5\times 5, 7\times 1, 1\times 7\}$. For the hyper-parameters, we set $\alpha=\beta=0.5, \tau_1=0.7, \tau_2=0.4$ in all experiments for a fair comparison with previous works \citep{05}, and set $\tau_3=0.40,\ \tau_4=0.55$. Moreover, $d=256, N_1=3$, $N_2=6$, $N_3=4$, $N=13$, and $N^{'}=48$. 

All experiments are implemented using PyTorch~\citep{28}, and trained on four NVIDIA Tesla V100 GPUs and tested on a single GPU. We conducted data augmentation, including random cropping, scaling, rotation, blurring, and flipping, same as the methods in \citep{06} and \citep{01} for both PASCAL-$5^i$ and COCO-$20^i$ datasets. The crop sizes are $473\times 473$ and $641\times 641$ for the PASCAL-$5^i$ dataset and the COCO-$20^i$ dataset, respectively.
\begin{sidewaystable*}[htbp]
    \caption{Comparison with 
state of the art methods in mIoU under 1-shot and 5-shot on the PASCAL-$5^i$ dataset. \textbf{Bold} denotes the best result and \underline{underline} denotes the second best result.}
    \centering
    \scalebox{0.8}{
    \begin{tabular}{l|c|ccccc|ccccc}
    \toprule
    \multirow{2}{*}{Method} & \multirow{2}{*}{Backbone} & \multicolumn{5}{c|}{1-shot} & \multicolumn{5}{c}{5-shot} \\
    \rule{0pt}{10pt}
    & & Fold-0 & Fold-1 & Fold-2 & Fold-3 & Mean & Fold-0 & Fold-1 & Fold-2 & Fold-3 & Mean\\
    \midrule
    PFENet \citep{06} {(TPAMI'20)} & VGG-16 & 56.90 & 68.20 & 54.40 & 52.40 & 58.00 & 59.00 & 69.10 & 54.80 &52.90 & 69.00  \\
    HSNet \citep{02} {(ICCV'21)} & VGG-16 & 59.60 & 65.70 & 59.60 & 54.00 & 59.70 & 64.90 & 69.00 & 64.10 & 58.60 & 64.10  \\
    MIANet \citep{01} {(CVPR'23)} & VGG-16 & 65.42 & 73.58 & 67.76 & 61.65 & 67.10 & 69.01 & 76.14 & 73.24 & 69.55 & \underline{71.99}  \\ 
    \midrule
    PFENet \citep{06} {(TPAMI'20)} & ResNet50 & 61.70 & 69.50 & 55.40 & 56.30 & 60.80 & 63.10 & 70.70 & 55.80 & 57.90 & 61.90   \\ 
    HSNet \citep{02} {(ICCV'21)} & ResNet50 & 64.30 & 70.70 & 60.30 & 60.50 & 64.00 & 70.30 & 73.20 & 67.40 & 67.10 & 69.50 \\
    DPCN \citep{07} {(CVPR'22)} & ResNet50 & 65.70 & 71.60 & \underline{69.10} & 60.60 & 66.70 & 70.00 & 73.20 & 70.90 & 65.50 & 69.90  \\
    NTRENet \citep{29} {(CVPR'22)} & ResNet50 & 65.40 & 72.30 & 59.40 & 59.80 & 64.20 & 66.20 & 72.80 & 61.70 & 62.20 & 65.70  \\
    SSP \citep{05} {(ECCV'22)} & ResNet50 & 60.50 & 67.80 & 66.40 & 51.00 & 61.40 & 67.50 & 72.30 & \textbf{75.20} & 62.10 & 69.30 \\
    CyCTR \citep{42} {(NeurIPS’21)} & ResNet50 & 65.70 & 71.00 & 59.50 & 59.70 & 64.00 & 69.30 & 73.50 & 63.80 & 63.50 & 67.50  \\
    BAM \citep{09} {(CVPR'22)} & ResNet50 & 68.97 & 73.59 & 67.55 & 61.13 & 67.81 & 70.59 & 75.05 & 70.79 & 67.20 & 70.91  \\
    HPA \citep{HPA} {(TPAMI'23)} & ResNet50 & 65.94 & 71.96 & 64.66& 56.78 & 64.84 & 70.54 & 73.28 & 68.37 & 63.41 & 68.90  \\
    FECANet \citep{04} {(TMM'23)} & ResNet50 & \underline{69.20} & 72.30 & 62.40 & \textbf{65.70} & 67.40 & {\textbf{72.90}} & 74.00 & 65.20 & 67.80 & 70.00  \\
    {QPENet \citep{43} (TMM'24)} & {ResNet50} & {65.20} & {71.90} & {64.10} & {59.50} & {65.20} & {68.40} & {74.0} & {67.40} & {65.20} & {68.80} \\ 
    MIANet \citep{01} {(CVPR'23)} & ResNet50 & 68.51 & \underline{75.76} & 67.46 & \underline{63.15} & \underline{68.72} & 70.20 & \textbf{77.38} & 70.02 & \textbf{68.77} & 71.59 \\ 
    \midrule
    Sym-Net (\textbf{Ours}) & ResNet50 & \textbf{69.91}&\textbf{76.16}&\textbf{ 69.59}& 62.96& \textbf{69.66} & {70.92} & \underline{77.17} & \underline{73.98} & \underline{68.30} & \textbf{72.59} \\
    \bottomrule
    \end{tabular}}
    \label{table1}
\end{sidewaystable*}
\begin{sidewaystable*}[htbp]
    \caption{Comparison with state of the art methods in mIoU under 1-shot and 5-shot on the COCO-$20^i$ dataset. \textbf{Bold} denotes the best result and \underline{underline} denotes the second best result.}
    \centering
    \scalebox{0.8}{
    \begin{tabular}{l|c|ccccc|ccccc}
    \toprule
    \multirow{2}{*}{Method} & \multirow{2}{*}{Backbone} & \multicolumn{5}{c|}{1-shot} & \multicolumn{5}{c}{5-shot} \\
    \rule{0pt}{10pt}
    & & Fold-0 & Fold-1 & Fold-2 & Fold-3 & Mean & Fold-0 & Fold-1 & Fold-2 & Fold-3 & Mean\\
    \midrule
    PFENet \citep{06} {(TPAMI'20)} & VGG-16 & 35.40 & 38.10 & 36.80 & 34.70 & 36.30 & 38.20 & 42.50 & 41.80 &38.90 & 40.40  \\
    MIANet \citep{01} {(CVPR'23)} & VGG-16 & 40.56 & 50.53 & 46.50 & 45.18 & 45.69 & 46.18 & 56.09 & \underline{52.33} & 49.54 & 51.03 \\ 
    \midrule
    PFENet \citep{06} {(TPAMI'20)} & ResNet50 & 36.50 & 38.60 & 34.50 & 33.80 & 35.80 & 36.50 & 43.30 & 37.80 & 38.40 & 39.00  \\ 
    HSNet \citep{02} {(ICCV'21)} & ResNet50 & 36.30 & 43.10 & 38.70 & 38.70 & 39.20 & 43.30 &51.30 & 48.20 & 45.00 & 46.90  \\
    DPCN \citep{07} {(CVPR'22)} & ResNet50 & 42.00 & 47.00 & 43.20 & 39.70 & 43.00 & 46.00 & 54.90 & 50.80 & 47.40 & 49.80  \\
    NTRENet \citep{29} {(CVPR'22)} & ResNet50 & 36.80 & 42.60 & 39.90 & 37.90 & 39.30 & 38.20 & 44.10 & 40.40 & 38.40 & 40.30 \\
    SSP \citep{05} {(ECCV'22)} & ResNet50 & 33.50 & 39.60 & 37.90 & 36.70 & 37.40 & 40.60 & 47.00 & 45.10 & 43.90 & 44.10  \\
    CyCTR \citep{42} {(NeurIPS’21)} & ResNet50 & 38.90 & 43.00 & 39.60 & 39.80 & 40.30 & 41.10 & 48.90 & 45.20 & 47.00 & 45.60 \\
    BAM \citep{09} {(CVPR'22)} & ResNet50 & \underline{43.41} & 50.59 & 47.49 & 43.42 & 46.23 & \textbf{49.26} & 54.20 & 51.63 & 49.55 & 51.16 \\
    HPA \citep{HPA} {(TPAMI'23)} & ResNet50 & 43.30 & 46.57 & 44.12 & 42.71 & 43.43 & 45.54 & 55.43 & 48.90 & 50.21 & 50.02 \\
    FECANet \citep{04} {(TMM'23)} & ResNet50 & 38.50	& 44.60 & 42.60 & 40.70 & 41.60 & 44.60 & 51.50 & 48.40 & 45.80 & 47.60 \\
    QPENet \citep{43} (TMM'24) & {ResNet50} & {41.50} & {47.30} & {40.90} & {39.40} & {42.30} & {47.30} & {52.40} & {44.30} & {44.90} & {47.20} \\
    MIANet \citep{01} {(CVPR'23)} & ResNet50 & 42.49 & \underline{52.95} & \underline{47.77} & \underline{47.42} & \underline{47.66} & 45.84 & \underline{58.18} & 51.29 & \underline{51.90} & \underline{51.65} \\ 
    \midrule
    Sym-Net (\textbf{Ours}) & ResNet50 &\textbf{43.63} & {\textbf{55.35}} & \textbf{50.32} &  \textbf{50.57} & \textbf{49.97} & \underline{47.46} & {\textbf{61.00}} &  \textbf{54.21} & \textbf{55.06}&\textbf{54.43}  \\
    \bottomrule
    \end{tabular}}
    \label{table2}
\end{sidewaystable*}
\begin{table}[t]
    \setlength{\tabcolsep}{5.5pt}
    \caption{Ablation studies on the effectiveness of three proposed modules. \textbf{Bold} indicates the best result.} 
    \centering
    \begin{tabular}{ccc|cccccc}
    \toprule
    SPM&APA&TDC & Fold-0 & Fold-1 & Fold-2 & Fold-3 & Mean&$\bigtriangleup$  \\ 
    \midrule 
    $\times$&$\times$&$\times$&64.14 & 72.91 & 65.39 &56.00 & 64.71&0.00\\
    \checkmark&$\times$&$\times$&66.31 & 72.99 & 66.28 &57.22 & 65.70&$\uparrow$0.99\\
    $\times$&$\times$&\checkmark&68.97 & 74.56 & 67.16 &60.92 & 67.90&$\uparrow$3.19\\
    \checkmark&\checkmark&$\times$&67.68&73.21& 66.69& 57.56& 66.29&$\uparrow$1.58\\
    \checkmark&$\times$&\checkmark&68.88&74.23& 68.56& 61.44& 68.28&$\uparrow$3.57\\
   \checkmark&\checkmark&\checkmark&\textbf{69.91}&\textbf{76.16}&\textbf{ 69.59}& \textbf{62.96}& \textbf{69.66}&$\uparrow$4.95\\
    \bottomrule
    \end{tabular}
    \footnotetext[*]{Note that APA is based on SPM, thus APA and SPM are bounded together. That is why APA is impossible to apply alone.}
    \label{table3}
\end{table}
\begin{table}[!t]
    \caption{Ablation studies on $\tau_3$ and $\tau_4$. \textbf{Bold} indicates the best result. The default configuration we choose are marked with $*$.} 
    \centering
    \begin{tabular}{ll|ccccc} 
    \toprule
    $\tau_3$&$\tau_4$ & Fold-0 & Fold-1 & Fold-2 & Fold-3 & Mean  \\ 
    \midrule
    0.45&0.55&70.04 & 75.79 & 68.67 &62.53 & 69.26\\
    0.45&0.60&\textbf{70.50} & 75.81 & 68.70 &62.74 & 69.38\\
    0.40$^*$&0.55$^*$&69.91 & \textbf{76.16} & \textbf{69.59} &\textbf{62.96} & \textbf{69.66}\\
    0.40&0.60&69.60&75.84& 68.31& 61.83& 68.90\\
    \bottomrule
    \end{tabular}
    \label{table5}
\end{table}
\subsection{Experimental Results} 
\subsubsection{Results on the PASCAL-\texorpdfstring{$5^i$} Dataset}
As shown in Table \ref{table1}, we report the comparison of mIoU results of our Sym-Net on the PASCAL-$5^i$ dataset under both 1-shot and 5-shot settings against other State-Of-The-Art (SOTA) methods. It can be seen that our method achieves the best results on both 1-shot and 5-shot settings with 69.66\% mIoU and 72.59\% mIoU, respectively. Furthermore, our method not only outperforms prototype-based baseline methods but also surpasses all correlation-based methods, such as HSNet \citep{02} and FECANet \citep{04}, suggesting the effectiveness of our proposed Sym-Net. 

To gain more insight, we provide qualitative comparisons of our method with the second best method MIANet on the PASCAL-$5^i$ dataset for various categories, as shown in Fig.~\ref{fig_3}. From this figure, we can see that even when there are substantial scale and spatial differences between the support and query images, our method can robustly predict the query object by relying on semantic information and local pixels in the query image. For instance, looking at the fourth and fifth columns in Fig.~\ref{fig_3}, our model can segment the query image very precisely, even when it has not observed the complete information of the support object. Additionally, our method enables us to handle multiple individual segments, as demonstrated in the fourth column. Sym-Net exhibits superior performance compared to MIANet, even in significant occlusion scenarios and hard cases, as illustrated in the last two columns. Moreover, we compare the visualization of the prior query masks generated by the non-parametric Hierarchical Prior Module (HPM) from MIANet~\citep{01} and our proposed SPM.
Note that we average four-scale masks from MIANet for a fair comparison. As shown in the third and fourth rows of Fig.~\ref{fig_3}, the prior masks that are generated by MIANet exhibit higher activation values in the background regions, which do not focus on the object and introduce more noisy information compared to our method. 

\subsubsection{Results on the COCO-\texorpdfstring{$20^i$} Dataset}
Table~\ref{table2} shows the comparison results on the more challenging COCO-$20^i$ dataset. Due to the inclusion of a larger number of object classes and more complex intra-class variations compared to the PASCAL-$5^i$ dataset, the performance of all methods is typically reduced. However, our approach can achieve the best results under both 1-shot and 5-shot settings with mIoU of 49.97\% and 54.43\%, respectively, 
revealing the effectiveness and robustness of Sym-Net in segmenting large-scale complex images. We attribute this superior performance to the robust capability of our method in mitigating intra-class discrepancies through the integration of query-fused prototype learning and injected semantic information.

\subsection{Ablation Study and Analysis}
We perform ablation studies on the PASCAL-$5^i$ dataset 
under 1-shot tasks with ResNet50 as the backbone network. 
These studies are conducted to verify the effectiveness of three proposed modules, explore the best configuration of the hyper-parameters $\tau_3$ and $\tau_4$, and investigate various visual-text aggregation methods, as well as different dimensions for the hyper-correlation features.\\

\begin{figure}[!t]
\centering
\includegraphics[width=1\linewidth]{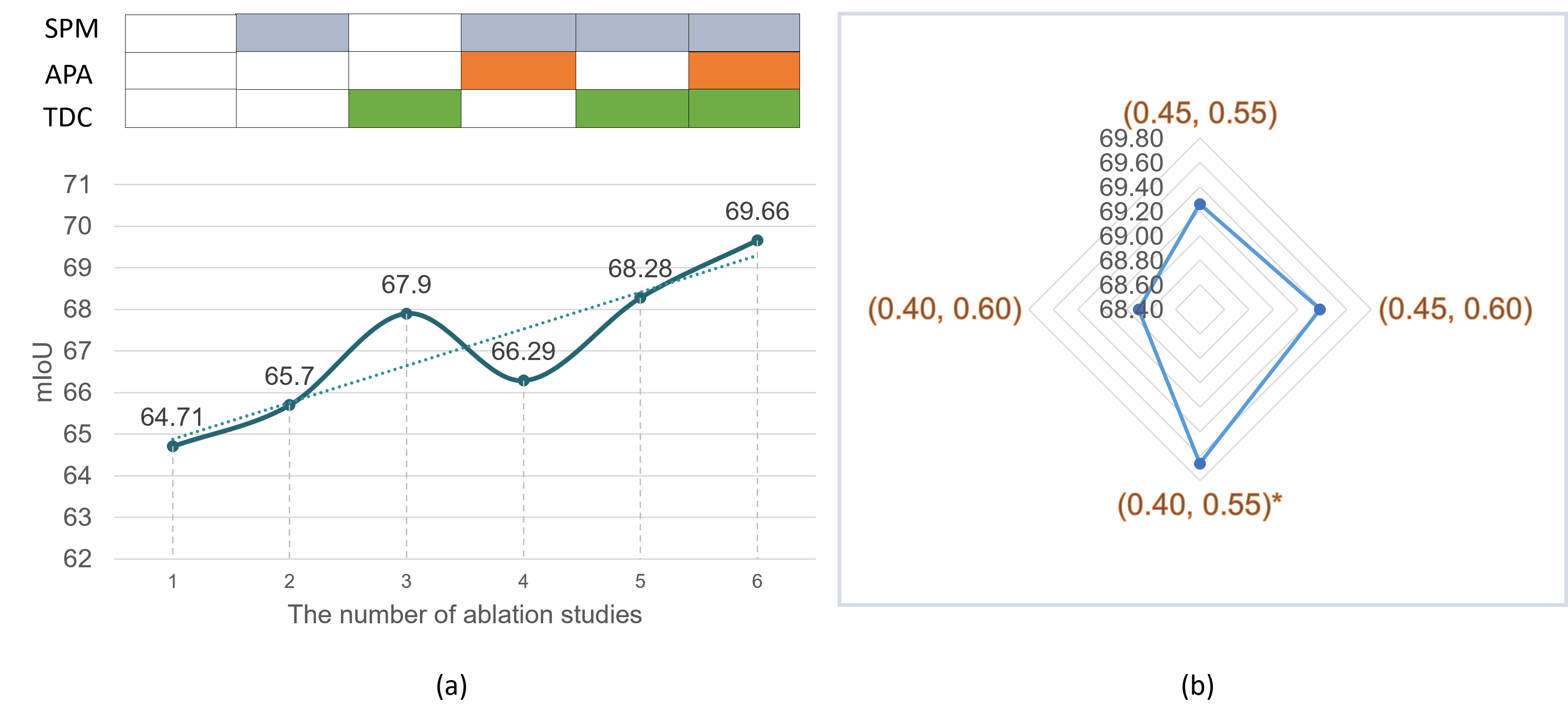}
\caption{The effect of three main modules and hyper-parameters $\tau_3$ and $\tau_4$ on the results of mIoU. (a) represents the effect of three main modules of Sym-Net, where the dashed line represents the trend and the colored rectangular boxes indicate the use of the corresponding modules. (b) shows the effect of $\tau_3$ and $\tau_4$ on mIoU, where the orange paired numbers are the value of $\tau_3$ and $\tau_4$.}
\label{fig8}
\end{figure}

\noindent{\textbf{Effectiveness of Three Proposed Modules.}}
We first utilize MAP \citep{41} as a baseline for prototype generation, and then incorporate our proposed APA, TDC, and SPM modules to enhance the baseline. Note that APA is based on SPM, thus APA and SPM are bound together. 
SPM strengthens the selection of the most relevant information from the query by using the self-activation kernel. The hybrid prototypes that are generated by APA possess class-specific information, enabling accurate classification of pixels similar to the target class during matching with query pixels. Therefore, APA and SPM enhance the model's ability to handle intra-class variations to some extent, and bring considerable improvements to the performance of the overall model, as shown in Table \ref{table3}. What's intriguing is that TDC is simple yet highly effective, achieving a 3.19\% increase in mIoU. For more clarification, we also visualize the trend of the effect with a line chart, as shown in Fig. \ref{fig8}(a). It can be seen that as we sequentially add each module, the results show a roughly linear increase.\\

\begin{figure}[!t]
\centering
\includegraphics[width=0.9\linewidth]{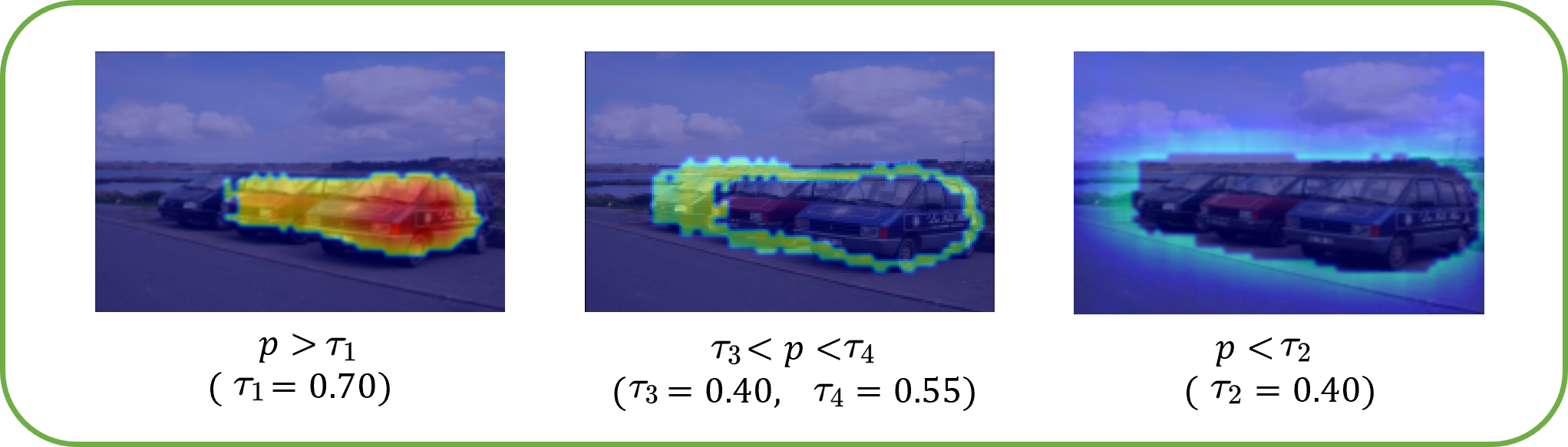}
\caption{Visualization of the operational mechanism of the hyper-parameters $\tau_1$, $\tau_2$, $\tau_3$ and $\tau_4$.}
\label{fig7}
\end{figure}

\noindent{\textbf{Hyper-parameter $\tau_3$ and $\tau_4$.}}
To explore the best configuration of the hyper-parameters $\tau_3$ and $\tau_4$, we conduct four-group ablation studies as shown in Table \ref{table5} and Fig. \ref{fig8}(b). Given that $\tau_3$ and $\tau_4$ delineate the boundary of the query object, with $\tau_3$ being the lower limit and $\tau_4$ the upper one, we vary $\tau_3$ from $\{0.40, 0.45\}$, and $\tau_4$ from $\{0.55, 0.60\}$. We find that a broad boundary range, such as $(\tau_3=0.40,\  \tau_4=0.60)$, results in an unclear and less defined boundary, contributing to a decrease in the model's performance. To elucidate how the hyper-parameters $\tau_1$, $\tau_2$, $\tau_3$ and $\tau_4$ partition the query image into three distinct regions, i.e., the high confidence query region, the ambiguous query region, and the background region, Fig.~\ref{fig7} visualizes the operational mechanism of these hyper-parameters in their default settings. The highlighted areas in each image represent the selected regions.\\

\noindent{\textbf{Different Prototype Aggregation Methods.}}
We investigate various visual-text alignment methods as shown in Table \ref{table4}. We can see that the dual-branch, the semantic information from text, and the joint triplet loss are almost equally significant in the APA module. 
\\

\noindent\textbf{Different Dimensions of Hyper-correlation Feature.}

\begin{table*}[!t]
\caption{Ablation studies on different prototype aggregation methods. \textbf{Bold} indicates the best result.\label{table4}}
\centering
\scalebox{0.9}{
    \begin{tabular}{l|ccccc} 
    \toprule
    Prototype aggregation methods & Fold-0 & Fold-1 & Fold-2 & Fold-3 & Mean  \\ 
    \midrule
    Support&68.88 & 74.23 & 68.56 &61.44 & 68.28\\
    Support \& Query&69.09 & 75.13 & 68.34 &61.97 & 68.63\\
    Support \& Word2Vec &69.45 & 75.10 & 68.81 &61.51 & 68.72\\
    Support \& Query \& Word2Vec w/o $\mathcal{L}_{co-triple}$&69.64&75.67& 68.64& 61.95& 68.98\\
    Support \& Query \& Word2Vec w/ $\mathcal{L}_{co-triple}$&\textbf{69.91}&\textbf{76.16}&\textbf{69.59}&\textbf{62.96}&\textbf{69.66}\\
    \bottomrule
    \end{tabular}
    }
\end{table*}
\begin{table}[!t]
\caption{Ablation studies on different dimensions of the hyper-correlation feature. \textbf{Bold} indicates the best result.\label{table7}}
\centering
    \begin{tabular}{c|ccccc} 
    \toprule
    Dimension of hyper-correlation feature & Fold-0 & Fold-1 & Fold-2 & Fold-3 & Mean  \\ 
    \midrule
    32&69.03 & 75.98 & 68.16 &61.84 & 68.75\\
    48 &69.91 & \textbf{76.16} & \textbf{69.59} &\textbf{62.96} & \textbf{69.66}\\
    64 &\textbf{70.22} & 75.35 & 68.49 &61.31 & 68.84\\
    \bottomrule
    \end{tabular}
\end{table}

As shown in Table \ref{table7}, we further conduct ablation studies on different dimensions of the hyper-correlation feature $\mathbf{F}^{hyper}$. Our model achieves the best results when the dimension of $\mathbf{F}^{hyper}$ is equal to 48 that is why we set it as the default choice. The results also indicate that a high dimensional  feature does not necessarily lead to superior outcomes.\\

\noindent\textbf{HPM vs. SPM.}
To validate the effectiveness of the proposed query mask generation module, i.e., SPM, we replace SPM with HPM proposed in MIANet.
The results of mIoU are $68.42\%$ (HPM) vs. $69.66\%$ (ours), which demonstrates the effectiveness of our prior mask generation method from the perspective of quantitative analysis. 

\begin{figure}[!t]
\centering
\includegraphics[width=1\linewidth]{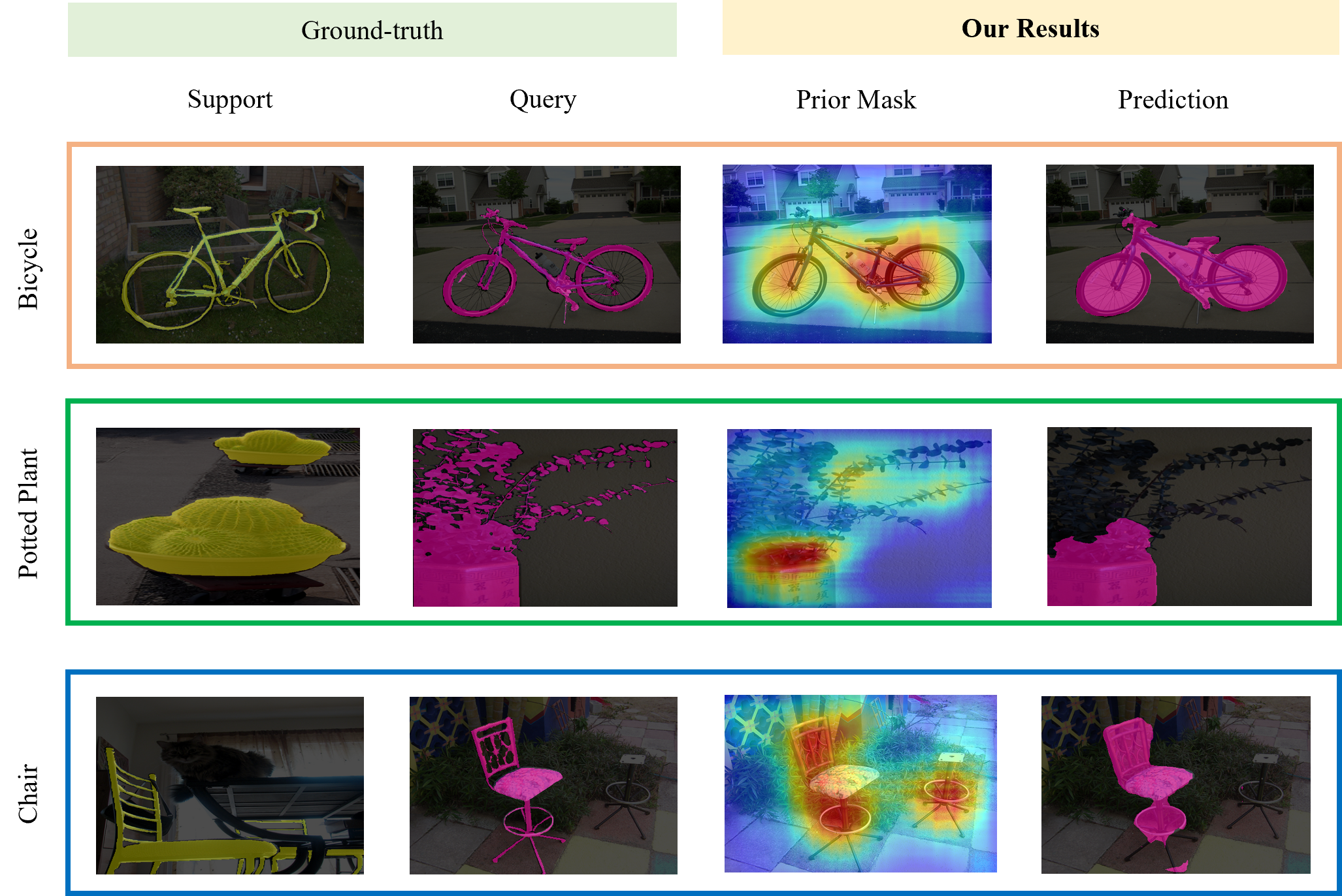}
\caption{Failure cases in segmentation. This can be attributed to their complex shapes and edge features, making it challenging for the model to accurately capture and represent these fine and intricate spatial structures. Additionally, the prior masks generated by the model in this scenario are not precise enough, thereby affecting the localization accuracy of local query pixels. Best viewed in color. }
\label{sup2}
\end{figure}
\subsection{Discussion and Future Work}
The proposed Sym-Net involves jointly learning support-query prototypes in a symmetrical manner to improve the performance of FSS. The symmetrical learning ensures that the learning process is balanced and consistent between the support and query samples. By leveraging mutual guidance and enforcing consistency, this approach achieves better generalization and accuracy in segmenting objects. However, our model exhibits shortcomings in handling cases with rich detailed features. We present some failure cases in Fig. \ref{sup2}. As depicted in these instances, sub-optimal outcomes are observed when predicting the leaves and neck of plants, as well as the skeletons of bicycles and chairs. This can be attributed to the intricate shapes and detailed edge features of these objects, which poses a challenge for the model to precisely capture and represent these fine spatial structures. Moreover, in this scenario, the prior masks generated by the model lack precision, consequently impacting the accuracy of localizing query pixels. However, in most cases as demonstrated in Fig. \ref{fig_3}, our model exhibits a strong capacity for segmenting unseen objects with just a few support samples.

As mentioned above, our model exhibits shortcomings in handling cases with rich detailed features. To overcome this limitation, our future work will involve investigating the integration of attention mechanisms with edge detection techniques. These strategies are designed to enhance the model's comprehension of challenging regions, thereby improving its capability to effectively handle complex shapes and intricate details.

\section{Conclusion}

To address the shortcomings of previous prototype-based Few-Shot Segmentation (FSS) methods in effectively mitigating intra-class discrepancy and preserving extensive multi-scale spatial information during prototype learning, we propose the novel Sym-Net, which demonstrates that jointly learning support-query prototypes in a symmetrical manner for FSS offers a promising direction to enhance segmentation performance with limited annotated data. Sym-Net alleviates intra-class discrepancy by employing a symmetrical query-support prototype representation and capturing multi-scale spatial dependencies through a Top-Down hyper-Correlation module (TDC). Our model demonstrates remarkable performance on both the PASCAL-$5^i$ and COCO-$20^i$ segmentation datasets, attributed to the effectiveness of hybrid prototype learning guided by a visual-text Alignment-based Prototype Aggregation module (APA) and a Self-activation-based Prior Mask generation module (SPM). The APA module adaptively integrates query-support pairwise and text embedding information to bridge intra-class discrepancy effectively. Moreover, the SPM module facilitates the extraction of more accurate object information, enabling parameter-free derivation of the query prototype through the generated prior mask. Experimental results validate the effectiveness of our model in tackling intra-class discrepancy and leveraging multi-scale spatial information in FSS. To handle complex shapes and intricate details more effectively, in the future, we will explore integrating edge information with attention mechanisms.




\bibliography{references, sn-bibliography}


\begin{thebibliography}{43}
\ifx \bisbn   \undefined \def \bisbn  #1{ISBN #1}\fi
\ifx \binits  \undefined \def \binits#1{#1}\fi
\ifx \bauthor  \undefined \def \bauthor#1{#1}\fi
\ifx \batitle  \undefined \def \batitle#1{#1}\fi
\ifx \bjtitle  \undefined \def \bjtitle#1{#1}\fi
\ifx \bvolume  \undefined \def \bvolume#1{\textbf{#1}}\fi
\ifx \byear  \undefined \def \byear#1{#1}\fi
\ifx \bissue  \undefined \def \bissue#1{#1}\fi
\ifx \bfpage  \undefined \def \bfpage#1{#1}\fi
\ifx \blpage  \undefined \def \blpage #1{#1}\fi
\ifx \burl  \undefined \def \burl#1{\textsf{#1}}\fi
\ifx \doiurl  \undefined \def \doiurl#1{\url{https://doi.org/#1}}\fi
\ifx \betal  \undefined \def \betal{\textit{et al.}}\fi
\ifx \binstitute  \undefined \def \binstitute#1{#1}\fi
\ifx \binstitutionaled  \undefined \def \binstitutionaled#1{#1}\fi
\ifx \bctitle  \undefined \def \bctitle#1{#1}\fi
\ifx \beditor  \undefined \def \beditor#1{#1}\fi
\ifx \bpublisher  \undefined \def \bpublisher#1{#1}\fi
\ifx \bbtitle  \undefined \def \bbtitle#1{#1}\fi
\ifx \bedition  \undefined \def \bedition#1{#1}\fi
\ifx \bseriesno  \undefined \def \bseriesno#1{#1}\fi
\ifx \blocation  \undefined \def \blocation#1{#1}\fi
\ifx \bsertitle  \undefined \def \bsertitle#1{#1}\fi
\ifx \bsnm \undefined \def \bsnm#1{#1}\fi
\ifx \bsuffix \undefined \def \bsuffix#1{#1}\fi
\ifx \bparticle \undefined \def \bparticle#1{#1}\fi
\ifx \barticle \undefined \def \barticle#1{#1}\fi
\bibcommenthead
\ifx \bconfdate \undefined \def \bconfdate #1{#1}\fi
\ifx \botherref \undefined \def \botherref #1{#1}\fi
\ifx \url \undefined \def \url#1{\textsf{#1}}\fi
\ifx \bchapter \undefined \def \bchapter#1{#1}\fi
\ifx \bbook \undefined \def \bbook#1{#1}\fi
\ifx \bcomment \undefined \def \bcomment#1{#1}\fi
\ifx \oauthor \undefined \def \oauthor#1{#1}\fi
\ifx \citeauthoryear \undefined \def \citeauthoryear#1{#1}\fi
\ifx \endbibitem  \undefined \def \endbibitem {}\fi
\ifx \bconflocation  \undefined \def \bconflocation#1{#1}\fi
\ifx \arxivurl  \undefined \def \arxivurl#1{\textsf{#1}}\fi
\csname PreBibitemsHook\endcsname

\bibitem[\protect\citeauthoryear{Tian et~al.}{2023}]{33}
\begin{bchapter}
\bauthor{\bsnm{Tian}, \binits{Z.}},
\bauthor{\bsnm{Cui}, \binits{J.}},
\bauthor{\bsnm{Jiang}, \binits{L.}},
\bauthor{\bsnm{Qi}, \binits{X.}},
\bauthor{\bsnm{Lai}, \binits{X.}},
\bauthor{\bsnm{Chen}, \binits{Y.}},
\bauthor{\bsnm{Liu}, \binits{S.}},
\bauthor{\bsnm{Jia}, \binits{J.}}:
\bctitle{Learning context-aware classifier for semantic segmentation}.
In: \bbtitle{Proceedings of the 37th AAAI Conference on Artificial Intelligence},
pp. \bfpage{2438}--\blpage{2446}
(\byear{2023})
\end{bchapter}
\endbibitem

\bibitem[\protect\citeauthoryear{Zhao et~al.}{2017}]{13}
\begin{bchapter}
\bauthor{\bsnm{Zhao}, \binits{H.}},
\bauthor{\bsnm{Shi}, \binits{J.}},
\bauthor{\bsnm{Qi}, \binits{X.}},
\bauthor{\bsnm{Wang}, \binits{X.}},
\bauthor{\bsnm{Jia}, \binits{J.}}:
\bctitle{Pyramid scene parsing network}.
In: \bbtitle{Proc. IEEE Conf. Comput. Vis. Pattern Recognit.},
pp. \bfpage{6230}--\blpage{6239}
(\byear{2017})
\end{bchapter}
\endbibitem

\bibitem[\protect\citeauthoryear{Wen et~al.}{2024 (Early Access)}]{44}
\begin{botherref}
\oauthor{\bsnm{Wen}, \binits{C.}},
\oauthor{\bsnm{Huang}, \binits{H.}},
\oauthor{\bsnm{Ma}, \binits{Y.}},
\oauthor{\bsnm{Yuan}, \binits{F.}},
\oauthor{\bsnm{Zhu}, \binits{H.}}:
Dual-guided frequency prototype network for few-shot semantic segmentation.
{IEEE} Trans. Multimedia,
1--15
(2024 (Early Access))
\doiurl{10.1109/TMM.2024.3383276}
\end{botherref}
\endbibitem

\bibitem[\protect\citeauthoryear{Tian et~al.}{2022}]{06}
\begin{barticle}
\bauthor{\bsnm{Tian}, \binits{Z.}},
\bauthor{\bsnm{Zhao}, \binits{H.}},
\bauthor{\bsnm{Shu}, \binits{M.}},
\bauthor{\bsnm{Yang}, \binits{Z.}},
\bauthor{\bsnm{Li}, \binits{R.}},
\bauthor{\bsnm{Jia}, \binits{J.}}:
\batitle{Prior guided feature enrichment network for few-shot segmentation}.
\bjtitle{{IEEE} Trans. Pattern Anal. Mach. Intell.}
\bvolume{44}(\bissue{2}),
\bfpage{1050}--\blpage{1065}
(\byear{2022})
\doiurl{10.1109/TPAMI.2020.3013717}
\end{barticle}
\endbibitem

\bibitem[\protect\citeauthoryear{Fan et~al.}{2022}]{05}
\begin{bchapter}
\bauthor{\bsnm{Fan}, \binits{Q.}},
\bauthor{\bsnm{Pei}, \binits{W.}},
\bauthor{\bsnm{Tai}, \binits{Y.-W.}},
\bauthor{\bsnm{Tang}, \binits{C.-K.}}:
\bctitle{Self-support few-shot semantic segmentation}.
In: \bbtitle{Proc. IEEE Int. Conf. Comput. Vis.},
pp. \bfpage{701}--\blpage{719}
(\byear{2022})
\end{bchapter}
\endbibitem

\bibitem[\protect\citeauthoryear{Liu et~al.}{2020}]{liu2020crnet}
\begin{bchapter}
\bauthor{\bsnm{Liu}, \binits{W.}},
\bauthor{\bsnm{Zhang}, \binits{C.}},
\bauthor{\bsnm{Lin}, \binits{G.}},
\bauthor{\bsnm{Liu}, \binits{F.}}:
\bctitle{{CRN}et: Cross-reference networks for few-shot segmentation}.
In: \bbtitle{Proc. IEEE Int. Conf. Comput. Vis.},
pp. \bfpage{4165}--\blpage{4173}
(\byear{2020})
\end{bchapter}
\endbibitem

\bibitem[\protect\citeauthoryear{Lang et~al.}{2023}]{10256677}
\begin{barticle}
\bauthor{\bsnm{Lang}, \binits{C.}},
\bauthor{\bsnm{Cheng}, \binits{G.}},
\bauthor{\bsnm{Tu}, \binits{B.}},
\bauthor{\bsnm{Li}, \binits{C.}},
\bauthor{\bsnm{Han}, \binits{J.}}:
\batitle{Retain and recover: Delving into information loss for few-shot segmentation}.
\bjtitle{IEEE Trans. Image Process}
\bvolume{32}(\bissue{1}),
\bfpage{5353}--\blpage{5365}
(\byear{2023})
\doiurl{10.1109/TIP.2023.3315555}
\end{barticle}
\endbibitem

\bibitem[\protect\citeauthoryear{Liu et~al.}{2022}]{CRCNet}
\begin{barticle}
\bauthor{\bsnm{Liu}, \binits{W.}},
\bauthor{\bsnm{Zhang}, \binits{C.}},
\bauthor{\bsnm{Lin}, \binits{G.}},
\bauthor{\bsnm{Liu}, \binits{F.}}:
\batitle{{CRCNet}: Few-shot segmentation with cross-reference and region–global conditional networks}.
\bjtitle{Int. J. Comput. Vis.}
\bvolume{130}(\bissue{12}),
\bfpage{1}--\blpage{18}
(\byear{2022})
\doiurl{10.1007/s11263-022-01677-7}
\end{barticle}
\endbibitem

\bibitem[\protect\citeauthoryear{Wang et~al.}{2019}]{wang2019panet}
\begin{bchapter}
\bauthor{\bsnm{Wang}, \binits{K.}},
\bauthor{\bsnm{Liew}, \binits{J.H.}},
\bauthor{\bsnm{Zou}, \binits{Y.}},
\bauthor{\bsnm{Zhou}, \binits{D.}},
\bauthor{\bsnm{Feng}, \binits{J.}}:
\bctitle{{PAN}et: Few-shot image semantic segmentation with prototype alignment}.
In: \bbtitle{Proc. IEEE Int. Conf. Comput. Vis.},
pp. \bfpage{9197}--\blpage{9206}
(\byear{2019})
\end{bchapter}
\endbibitem

\bibitem[\protect\citeauthoryear{Min et~al.}{2021}]{02}
\begin{bchapter}
\bauthor{\bsnm{Min}, \binits{J.}},
\bauthor{\bsnm{Kang}, \binits{D.}},
\bauthor{\bsnm{Cho}, \binits{M.}}:
\bctitle{Hypercorrelation squeeze for few-shot segmenation}.
In: \bbtitle{Proc. IEEE Int. Conf. Comput. Vis.},
pp. \bfpage{6921}--\blpage{6932}
(\byear{2021})
\end{bchapter}
\endbibitem

\bibitem[\protect\citeauthoryear{Chen et~al.}{2018}]{23}
\begin{barticle}
\bauthor{\bsnm{Chen}, \binits{L.-C.}},
\bauthor{\bsnm{Papandreou}, \binits{G.}},
\bauthor{\bsnm{Kokkinos}, \binits{I.}},
\bauthor{\bsnm{Murphy}, \binits{K.}},
\bauthor{\bsnm{Yuille}, \binits{A.L.}}:
\batitle{Deep{L}ab: Semantic image segmentation with deep convolutional nets, atrous convolution, and fully connected {CRF}s}.
\bjtitle{{IEEE} Trans. Pattern Anal. Mach. Intell.}
\bvolume{40}(\bissue{4}),
\bfpage{834}--\blpage{848}
(\byear{2018})
\doiurl{10.1109/TPAMI.2017.2699184}
\end{barticle}
\endbibitem

\bibitem[\protect\citeauthoryear{Long et~al.}{2015}]{32}
\begin{bchapter}
\bauthor{\bsnm{Long}, \binits{J.}},
\bauthor{\bsnm{Shelhamer}, \binits{E.}},
\bauthor{\bsnm{Darrell}, \binits{T.}}:
\bctitle{Fully convolutional networks for semantic segmentation}.
In: \bbtitle{Proc. IEEE Conf. Comput. Vis. Pattern Recognit.},
pp. \bfpage{3431}--\blpage{3440}
(\byear{2015})
\end{bchapter}
\endbibitem

\bibitem[\protect\citeauthoryear{Huang et~al.}{2023}]{24}
\begin{barticle}
\bauthor{\bsnm{Huang}, \binits{Z.}},
\bauthor{\bsnm{Wang}, \binits{X.}},
\bauthor{\bsnm{Wei}, \binits{Y.}},
\bauthor{\bsnm{Huang}, \binits{L.}},
\bauthor{\bsnm{Shi}, \binits{H.}},
\bauthor{\bsnm{Liu}, \binits{W.}},
\bauthor{\bsnm{Huang}, \binits{T.S.}}:
\batitle{{CCNet}: Criss-cross attention for semantic segmentation}.
\bjtitle{{IEEE} Trans. Pattern Anal. Mach. Intell.}
\bvolume{45}(\bissue{6}),
\bfpage{6896}--\blpage{6908}
(\byear{2023})
\doiurl{10.1109/ICCV.2019.00069}
\end{barticle}
\endbibitem

\bibitem[\protect\citeauthoryear{Xie et~al.}{2021}]{25}
\begin{bchapter}
\bauthor{\bsnm{Xie}, \binits{E.}},
\bauthor{\bsnm{Wang}, \binits{W.}},
\bauthor{\bsnm{Yu}, \binits{Z.}},
\bauthor{\bsnm{Anandkumar}, \binits{A.}},
\bauthor{\bsnm{Alvarez}, \binits{J.M.}},
\bauthor{\bsnm{Luo}, \binits{P.}}:
\bctitle{{SegFormer}: Simple and efficient design for semantic segmentation with transformers}.
In: \bbtitle{Proc. Adv. Neural Inf. Process. Syst.},
pp. \bfpage{12077}--\blpage{12090}
(\byear{2021})
\end{bchapter}
\endbibitem

\bibitem[\protect\citeauthoryear{Zheng et~al.}{2021}]{26}
\begin{bchapter}
\bauthor{\bsnm{Zheng}, \binits{S.}},
\bauthor{\bsnm{Lu}, \binits{J.}},
\bauthor{\bsnm{Zhao}, \binits{H.}},
\bauthor{\bsnm{Zhu}, \binits{X.}},
\bauthor{\bsnm{Luo}, \binits{Z.}},
\bauthor{\bsnm{Wang}, \binits{Y.}},
\bauthor{\bsnm{Fu}, \binits{Y.}},
\bauthor{\bsnm{Feng}, \binits{J.}},
\bauthor{\bsnm{Xiang}, \binits{T.}},
\bauthor{\bsnm{Torr}, \binits{P.H.S.}},
\bauthor{\bsnm{Zhang}, \binits{L.}}:
\bctitle{Rethinking semantic segmentation from a sequence-to-sequence perspective with transformers}.
In: \bbtitle{Proc. IEEE Conf. Comput. Vis. Pattern Recognit.},
pp. \bfpage{6877}--\blpage{6886}
(\byear{2021})
\end{bchapter}
\endbibitem

\bibitem[\protect\citeauthoryear{Kirillov et~al.}{2023}]{kirillov2023segany}
\begin{bchapter}
\bauthor{\bsnm{Kirillov}, \binits{A.}},
\bauthor{\bsnm{Mintun}, \binits{E.}},
\bauthor{\bsnm{Ravi}, \binits{N.}},
\bauthor{\bsnm{Mao}, \binits{H.}},
\bauthor{\bsnm{Rolland}, \binits{C.}},
\bauthor{\bsnm{Gustafson}, \binits{L.}},
\bauthor{\bsnm{Xiao}, \binits{T.}},
\bauthor{\bsnm{Whitehead}, \binits{S.}},
\bauthor{\bsnm{Berg}, \binits{A.C.}},
\bauthor{\bsnm{Lo}, \binits{W.-Y.}},
\bauthor{\bsnm{Dollar}, \binits{P.}},
\bauthor{\bsnm{Girshick}, \binits{R.}}:
\bctitle{Segment anything}.
In: \bbtitle{Proc. IEEE Int. Conf. Comput. Vis.},
pp. \bfpage{4015}--\blpage{4026}
(\byear{2023})
\end{bchapter}
\endbibitem

\bibitem[\protect\citeauthoryear{Vinyals et~al.}{2016}]{31}
\begin{bchapter}
\bauthor{\bsnm{Vinyals}, \binits{O.}},
\bauthor{\bsnm{Blundell}, \binits{C.}},
\bauthor{\bsnm{Lillicrap}, \binits{T.}},
\bauthor{\bsnm{Kavukcuoglu}, \binits{K.}},
\bauthor{\bsnm{Wierstra}, \binits{D.}}:
\bctitle{Matching networks for one shot learning}.
In: \bbtitle{Proc. Adv. Neural Inf. Process. Syst.},
pp. \bfpage{3637}--\blpage{3645}
(\byear{2016})
\end{bchapter}
\endbibitem

\bibitem[\protect\citeauthoryear{Snell et~al.}{2017}]{39}
\begin{bchapter}
\bauthor{\bsnm{Snell}, \binits{J.}},
\bauthor{\bsnm{Swersky}, \binits{K.}},
\bauthor{\bsnm{Zemel}, \binits{R.}}:
\bctitle{Prototypical networks for few-shot learning}.
In: \bbtitle{Proc. Adv. Neural Inf. Process. Syst.},
pp. \bfpage{4080}--\blpage{4090}
(\byear{2017})
\end{bchapter}
\endbibitem

\bibitem[\protect\citeauthoryear{Hospedales et~al.}{2022}]{40}
\begin{barticle}
\bauthor{\bsnm{Hospedales}, \binits{T.}},
\bauthor{\bsnm{Antoniou}, \binits{A.}},
\bauthor{\bsnm{Micaelli}, \binits{P.}},
\bauthor{\bsnm{Storkey}, \binits{A.}}:
\batitle{Meta-learning in neural networks: A survey}.
\bjtitle{{IEEE} Trans. Pattern Anal. Mach. Intell.}
\bvolume{44}(\bissue{9}),
\bfpage{5149}--\blpage{5169}
(\byear{2022})
\doiurl{10.1109/TPAMI.2021.3079209}
\end{barticle}
\endbibitem

\bibitem[\protect\citeauthoryear{Dong and Xing}{2019}]{34}
\begin{bchapter}
\bauthor{\bsnm{Dong}, \binits{N.}},
\bauthor{\bsnm{Xing}, \binits{E.P.}}:
\bctitle{Few-shot semantic segmentation with prototype learning}.
In: \bbtitle{Proc. Brit. Mach. Vis. Conf.},
pp. \bfpage{1}--\blpage{13}
(\byear{2019})
\end{bchapter}
\endbibitem

\bibitem[\protect\citeauthoryear{Liu et~al.}{2022}]{29}
\begin{bchapter}
\bauthor{\bsnm{Liu}, \binits{Y.}},
\bauthor{\bsnm{Liu}, \binits{N.}},
\bauthor{\bsnm{Cao}, \binits{Q.}},
\bauthor{\bsnm{Yao}, \binits{X.}},
\bauthor{\bsnm{Han}, \binits{J.}},
\bauthor{\bsnm{Shao}, \binits{L.}}:
\bctitle{Learning non-target knowledge for few-shot semantic segmentation}.
In: \bbtitle{Proc. IEEE Conf. Comput. Vis. Pattern Recognit.},
pp. \bfpage{11563}--\blpage{11572}
(\byear{2022})
\end{bchapter}
\endbibitem

\bibitem[\protect\citeauthoryear{Yang et~al.}{2023}]{01}
\begin{bchapter}
\bauthor{\bsnm{Yang}, \binits{Y.}},
\bauthor{\bsnm{Chen}, \binits{Q.}},
\bauthor{\bsnm{Feng}, \binits{Y.}},
\bauthor{\bsnm{Huang}, \binits{T.}}:
\bctitle{{MIANet}: Aggregating unbiased instance and general information for few-shot semantic segmentation}.
In: \bbtitle{Proc. IEEE Conf. Comput. Vis. Pattern Recognit.},
pp. \bfpage{7131}--\blpage{7140}
(\byear{2023})
\end{bchapter}
\endbibitem

\bibitem[\protect\citeauthoryear{Yang et~al.}{2020}]{22}
\begin{bchapter}
\bauthor{\bsnm{Yang}, \binits{B.}},
\bauthor{\bsnm{Liu}, \binits{C.}},
\bauthor{\bsnm{Li}, \binits{B.}},
\bauthor{\bsnm{Jiao}, \binits{J.}},
\bauthor{\bsnm{Ye}, \binits{Q.}}:
\bctitle{Prototype mixture models for few-shot semantic segmentation}.
In: \bbtitle{Proc. Eur. Conf. Comput. Vis.},
pp. \bfpage{763}--\blpage{778}
(\byear{2020})
\end{bchapter}
\endbibitem

\bibitem[\protect\citeauthoryear{Zhang et~al.}{2020}]{41}
\begin{barticle}
\bauthor{\bsnm{Zhang}, \binits{X.}},
\bauthor{\bsnm{Wei}, \binits{Y.}},
\bauthor{\bsnm{Yang}, \binits{Y.}},
\bauthor{\bsnm{Huang}, \binits{T.S.}}:
\batitle{{SG-O}ne: Similarity guidance network for one-shot semantic segmentation}.
\bjtitle{IEEE Trans. Cybernetics}
\bvolume{50}(\bissue{9}),
\bfpage{3855}--\blpage{3865}
(\byear{2020})
\doiurl{10.1109/TCYB.2020.2992433}
\end{barticle}
\endbibitem

\bibitem[\protect\citeauthoryear{Peng et~al.}{2023}]{10}
\begin{bchapter}
\bauthor{\bsnm{Peng}, \binits{B.}},
\bauthor{\bsnm{Tian}, \binits{Z.}},
\bauthor{\bsnm{Wu}, \binits{X.}},
\bauthor{\bsnm{Wang}, \binits{C.}},
\bauthor{\bsnm{Liu}, \binits{S.}},
\bauthor{\bsnm{Su}, \binits{J.}},
\bauthor{\bsnm{Jia}, \binits{J.}}:
\bctitle{Hierarchical dense correlation distillation for few-shot segmentation}.
In: \bbtitle{Proc. IEEE Conf. Comput. Vis. Pattern Recognit.},
pp. \bfpage{1}--\blpage{16}
(\byear{2023})
\end{bchapter}
\endbibitem

\bibitem[\protect\citeauthoryear{Nie et~al.}{2024}]{nie2024crossdomain}
\begin{bchapter}
\bauthor{\bsnm{Nie}, \binits{J.}},
\bauthor{\bsnm{Xing}, \binits{Y.}},
\bauthor{\bsnm{Zhang}, \binits{G.}},
\bauthor{\bsnm{Yan}, \binits{P.}},
\bauthor{\bsnm{Xiao}, \binits{A.}},
\bauthor{\bsnm{Tan}, \binits{Y.-P.}},
\bauthor{\bsnm{Kot}, \binits{A.C.}},
\bauthor{\bsnm{Lu}, \binits{S.}}:
\bctitle{Cross-domain few-shot segmentation via iterative support-query correspondence mining}.
In: \bbtitle{Proc. IEEE Int. Conf. Comput. Vis.},
pp. \bfpage{1}--\blpage{11}
(\byear{2024})
\end{bchapter}
\endbibitem

\bibitem[\protect\citeauthoryear{Zhu et~al.}{2024}]{zhu2024llafs}
\begin{bchapter}
\bauthor{\bsnm{Zhu}, \binits{L.}},
\bauthor{\bsnm{Chen}, \binits{T.}},
\bauthor{\bsnm{Ji}, \binits{D.}},
\bauthor{\bsnm{Ye}, \binits{J.}},
\bauthor{\bsnm{Liu}, \binits{J.}}:
\bctitle{{LLaFS}: When large language models meet few-shot segmentation}.
In: \bbtitle{Proc. IEEE Int. Conf. Comput. Vis.},
pp. \bfpage{1}--\blpage{11}
(\byear{2024})
\end{bchapter}
\endbibitem

\bibitem[\protect\citeauthoryear{Lüddecke and Ecker}{2022}]{20}
\begin{bchapter}
\bauthor{\bsnm{Lüddecke}, \binits{T.}},
\bauthor{\bsnm{Ecker}, \binits{A.}}:
\bctitle{Image segmentation using text and image prompts}.
In: \bbtitle{Proc. IEEE Conf. Comput. Vis.},
pp. \bfpage{7076}--\blpage{7086}
(\byear{2022})
\end{bchapter}
\endbibitem

\bibitem[\protect\citeauthoryear{Radford et~al.}{2021}]{27}
\begin{bchapter}
\bauthor{\bsnm{Radford}, \binits{A.}},
\bauthor{\bsnm{Kim}, \binits{J.W.}},
\bauthor{\bsnm{Hallacy}, \binits{C.}},
\bauthor{\bsnm{Ramesh}, \binits{A.}},
\bauthor{\bsnm{Goh}, \binits{G.}},
\bauthor{\bsnm{Agarwal}, \binits{S.}},
\bauthor{\bsnm{Sastry}, \binits{G.}},
\bauthor{\bsnm{Askell}, \binits{A.}},
\bauthor{\bsnm{Mishkin}, \binits{P.}},
\bauthor{\bsnm{Clark}, \binits{J.}},
\bauthor{\bsnm{Krueger}, \binits{G.}},
\bauthor{\bsnm{Sutskever}, \binits{I.}}:
\bctitle{Learning transferable visual models from natural language supervision}.
In: \bbtitle{Proc. Mach. Learn. Res.},
pp. \bfpage{8748}--\blpage{8763}
(\byear{2021})
\end{bchapter}
\endbibitem

\bibitem[\protect\citeauthoryear{Wang et~al.}{2022}]{21}
\begin{bchapter}
\bauthor{\bsnm{Wang}, \binits{H.}},
\bauthor{\bsnm{Liu}, \binits{L.}},
\bauthor{\bsnm{Zhang}, \binits{W.}},
\bauthor{\bsnm{Zhang}, \binits{J.}},
\bauthor{\bsnm{Gan}, \binits{Z.}},
\bauthor{\bsnm{Wang}, \binits{Y.}},
\bauthor{\bsnm{Wang}, \binits{C.}},
\bauthor{\bsnm{Wang}, \binits{H.}}:
\bctitle{Iterative few-shot semantic segmentation from image label text}.
In: \bbtitle{Int. Joint Conf. Artif. Intell.},
pp. \bfpage{1385}--\blpage{1392}
(\byear{2022})
\end{bchapter}
\endbibitem

\bibitem[\protect\citeauthoryear{Shaban et~al.}{2017}]{35}
\begin{bchapter}
\bauthor{\bsnm{Shaban}, \binits{A.}},
\bauthor{\bsnm{Bansal}, \binits{S.}},
\bauthor{\bsnm{Liu}, \binits{Z.}},
\bauthor{\bsnm{Essa}, \binits{I.}},
\bauthor{\bsnm{Boots}, \binits{B.}}:
\bctitle{One-shot learning for semantic segmentation}.
In: \bbtitle{Proc. Brit. Mach. Vis. Conf.},
pp. \bfpage{1}--\blpage{13}
(\byear{2017})
\end{bchapter}
\endbibitem

\bibitem[\protect\citeauthoryear{Everingham et~al.}{2010}]{14}
\begin{barticle}
\bauthor{\bsnm{Everingham}, \binits{M.}},
\bauthor{\bsnm{Gool}, \binits{L.V.}},
\bauthor{\bsnm{Williams}, \binits{C.K.I.}},
\bauthor{\bsnm{Winn}, \binits{J.}},
\bauthor{\bsnm{Zisserman}, \binits{A.}}:
\batitle{The pascal visual object classes ({VOC}) challenge}.
\bjtitle{Int. J. Comput. Vis.}
\bvolume{88}(\bissue{2}),
\bfpage{303}--\blpage{338}
(\byear{2010})
\doiurl{10.1007/s11263-009-0275-4}
\end{barticle}
\endbibitem

\bibitem[\protect\citeauthoryear{Nguyen and Todorovic}{2019}]{16}
\begin{bchapter}
\bauthor{\bsnm{Nguyen}, \binits{K.}},
\bauthor{\bsnm{Todorovic}, \binits{S.}}:
\bctitle{Feature weighting and boosting for few-shot segmentation}.
In: \bbtitle{Proc. IEEE Int. Conf. Comput. Vis.},
pp. \bfpage{622}--\blpage{631}
(\byear{2019})
\end{bchapter}
\endbibitem

\bibitem[\protect\citeauthoryear{Lin et~al.}{2014}]{17}
\begin{bchapter}
\bauthor{\bsnm{Lin}, \binits{T.-Y.}},
\bauthor{\bsnm{Maire}, \binits{M.}},
\bauthor{\bsnm{Belongie}, \binits{S.}},
\bauthor{\bsnm{Hays}, \binits{J.}},
\bauthor{\bsnm{Perona}, \binits{P.}},
\bauthor{\bsnm{Ramanan}, \binits{D.}},
\bauthor{\bsnm{Doll{\'a}r}, \binits{P.}},
\bauthor{\bsnm{Zitnick}, \binits{C.L.}}:
\bctitle{Microsoft coco: Common objects in context}.
In: \bbtitle{Proc. Eur. Conf. Comput. Vis.},
pp. \bfpage{740}--\blpage{755}
(\byear{2014})
\end{bchapter}
\endbibitem

\bibitem[\protect\citeauthoryear{He et~al.}{2016}]{11}
\begin{bchapter}
\bauthor{\bsnm{He}, \binits{K.}},
\bauthor{\bsnm{Zhang}, \binits{X.}},
\bauthor{\bsnm{Ren}, \binits{S.}},
\bauthor{\bsnm{Sun}, \binits{J.}}:
\bctitle{Deep residual learning for image recognition}.
In: \bbtitle{Proc. IEEE Conf. Comput. Vis. Pattern Recognit.},
pp. \bfpage{770}--\blpage{778}
(\byear{2016})
\end{bchapter}
\endbibitem

\bibitem[\protect\citeauthoryear{Mikolov et~al.}{2013}]{word2vec}
\begin{bchapter}
\bauthor{\bsnm{Mikolov}, \binits{T.}},
\bauthor{\bsnm{Sutskever}, \binits{I.}},
\bauthor{\bsnm{Chen}, \binits{K.}},
\bauthor{\bsnm{Corrado}, \binits{G.}},
\bauthor{\bsnm{Dean}, \binits{J.}}:
\bctitle{Distributed representations ofwords and phrases and their compositionality}.
In: \bbtitle{Proc. Adv. Neural Inf. Process. Syst.},
pp. \bfpage{1}--\blpage{9}
(\byear{2013})
\end{bchapter}
\endbibitem

\bibitem[\protect\citeauthoryear{Lang et~al.}{2022}]{09}
\begin{bchapter}
\bauthor{\bsnm{Lang}, \binits{C.}},
\bauthor{\bsnm{Cheng}, \binits{G.}},
\bauthor{\bsnm{Tu}, \binits{B.}},
\bauthor{\bsnm{Han}, \binits{J.}}:
\bctitle{Learning what not to segment: A new perspective on few-shot segmentation}.
In: \bbtitle{Proc. IEEE Conf. Comput. Vis. Pattern Recognit.},
pp. \bfpage{1}--\blpage{12}
(\byear{2022})
\end{bchapter}
\endbibitem

\bibitem[\protect\citeauthoryear{Paszke et~al.}{2019}]{28}
\begin{bchapter}
\bauthor{\bsnm{Paszke}, \binits{A.}},
\bauthor{\bsnm{Gross}, \binits{S.}},
\bauthor{\bsnm{Massa}, \binits{F.}},
\bauthor{\bsnm{al}}:
\bctitle{{PyTorch}: An imperative style, high-performance deep learning library}.
In: \bbtitle{Proc. Adv. Neural Inf. Process. Syst.},
pp. \bfpage{1}--\blpage{12}
(\byear{2019})
\end{bchapter}
\endbibitem

\bibitem[\protect\citeauthoryear{Liu et~al.}{2022}]{07}
\begin{bchapter}
\bauthor{\bsnm{Liu}, \binits{J.}},
\bauthor{\bsnm{Bao}, \binits{Y.}},
\bauthor{\bsnm{Xie}, \binits{G.-S.}},
\bauthor{\bsnm{Xiong}, \binits{H.}},
\bauthor{\bsnm{Sonke}, \binits{J.-J.}},
\bauthor{\bsnm{Gavves}, \binits{E.}}:
\bctitle{Dynamic prototype convolution network for few-shot semantic segmentation}.
In: \bbtitle{Proc. IEEE Conf. Comput. Vis. Pattern Recognit.},
pp. \bfpage{11543}--\blpage{11552}
(\byear{2022})
\end{bchapter}
\endbibitem

\bibitem[\protect\citeauthoryear{Zhang et~al.}{2021}]{42}
\begin{bchapter}
\bauthor{\bsnm{Zhang}, \binits{G.}},
\bauthor{\bsnm{Kang}, \binits{G.}},
\bauthor{\bsnm{Yang}, \binits{Y.}},
\bauthor{\bsnm{Wei}, \binits{Y.}}:
\bctitle{Few-shot segmentation via cycle-consistent transformer}.
In: \bbtitle{Proc. Adv. Neural Inf. Process. Syst.},
pp. \bfpage{21984}--\blpage{21996}
(\byear{2021})
\end{bchapter}
\endbibitem

\bibitem[\protect\citeauthoryear{Cheng et~al.}{2023}]{HPA}
\begin{barticle}
\bauthor{\bsnm{Cheng}, \binits{G.}},
\bauthor{\bsnm{Lang}, \binits{C.}},
\bauthor{\bsnm{Han}, \binits{J.}}:
\batitle{Holistic prototype activation for few-shot segmentation}.
\bjtitle{{IEEE} Trans. Pattern Anal. Mach. Intell.}
\bvolume{45}(\bissue{4}),
\bfpage{4650}--\blpage{4666}
(\byear{2023})
\doiurl{10.1109/TPAMI.2022.3193587}
\end{barticle}
\endbibitem

\bibitem[\protect\citeauthoryear{Liu et~al.}{2023}]{04}
\begin{barticle}
\bauthor{\bsnm{Liu}, \binits{H.}},
\bauthor{\bsnm{Peng}, \binits{P.}},
\bauthor{\bsnm{Chen}, \binits{T.}},
\bauthor{\bsnm{al}}:
\batitle{{FECANet}: Boosting few-shot semantic segmentation with feature-enhanced context-aware network}.
\bjtitle{{IEEE} Trans. Multimedia}
\bvolume{25}(\bissue{1}),
\bfpage{8580}--\blpage{8592}
(\byear{2023})
\doiurl{10.1109/TMM.2023.3238521}
\end{barticle}
\endbibitem

\bibitem[\protect\citeauthoryear{Cong et~al.}{2024}]{43}
\begin{barticle}
\bauthor{\bsnm{Cong}, \binits{R.}},
\bauthor{\bsnm{Xiong}, \binits{H.}},
\bauthor{\bsnm{Chen}, \binits{J.}},
\bauthor{\bsnm{Zhang}, \binits{W.}},
\bauthor{\bsnm{Huang}, \binits{Q.}},
\bauthor{\bsnm{Zhao}, \binits{Y.}}:
\batitle{Query-guided prototype evolution network for few-shot segmentation}.
\bjtitle{{IEEE} Trans. Multimedia}
\bvolume{26}(\bissue{1}),
\bfpage{6501}--\blpage{6512}
(\byear{2024})
\doiurl{10.1109/TMM.2024.3352921}
\end{barticle}
\endbibitem

\end{thebibliography}

\end{document}